\documentclass{article} %
\usepackage[table]{xcolor}  %
\usepackage[final]{colm2026_conference}

\usepackage{microtype}
\usepackage{hyperref}
\usepackage{url}
\usepackage{booktabs}

\usepackage{lineno}

\definecolor{darkblue}{rgb}{0, 0, 0.5}
\hypersetup{colorlinks=true, citecolor=darkblue, linkcolor=darkblue, urlcolor=darkblue}

\title{Inference-Time Scaling of Diffusion Language Models \\via Trajectory Refinement}

\author{%
  Meihua Dang \\
  Stanford University \\
  \texttt{mhdang@cs.stanford.edu} \\
  \And Jiaqi Han \\
  Stanford University \\
  \texttt{jiaqihan@cs.stanford.edu} \\
  \And Minkai Xu \\
  Stanford University \\
  \texttt{minkai@cs.stanford.edu} \\
  \AND
  Kai Xu \\
    Red Hat AI Innovation \\
  \texttt{xuk@redhat.com} \\
  \And Akash Srivastava\\
  Red Hat AI Innovation \\
  \texttt{akash@redhat.com} \\
  \And
  Stefano Ermon \\
  Stanford University \\
  \texttt{ermon@cs.stanford.edu} \\ 
}

\usepackage{colortbl}       %
\usepackage{enumitem}

\usepackage{amsmath,amsfonts,bm}

\def\eqref#1{equation~\ref{#1}}

\def\1{\bm{1}}

\def\rvc{{\mathbf{c}}}

\def\rvm{{\mathbf{m}}}

\def\rvx{{\mathbf{x}}}

\DeclareMathAlphabet{\mathsfit}{\encodingdefault}{\sfdefault}{m}{sl}
\SetMathAlphabet{\mathsfit}{bold}{\encodingdefault}{\sfdefault}{bx}{n}

\newcommand{\E}{\mathbb{E}}

\newcommand{\KL}{D_{\mathrm{KL}}}

\DeclareMathOperator*{\argmax}{arg\,max}

\def\x0{\rvx_0}

\newcommand{\chg}[1]{{#1}}
\definecolor{darkgreen}{rgb}{0.0,0.5,0.0}
\definecolor{darkblue}{rgb}{0.0,0.2,0.7}
\usepackage{multirow}
\usepackage{subcaption}
\usepackage{nicematrix}
\usepackage{wrapfig}
\usepackage{booktabs}
\usepackage{array}
\usepackage{amsthm}
\usepackage{algorithm}
\usepackage{algpseudocode}
\usepackage{makecell}
\usepackage{calc}
\newcolumntype{P}[1]{>{\centering\arraybackslash}p{#1}}
\newcolumntype{L}[1]{>{\raggedright\arraybackslash}p{#1}}
\newcommand{\method}{PG-DLM}
\newcommand{\myparagraph}[1]{\paragraph{#1}}
\newcommand\given[1][]{\:#1\vert\:}

\theoremstyle{plain}
\newtheorem{theorem}{Theorem}[section]

\theoremstyle{definition}

\theoremstyle{plain}
\newtheorem{remark}[theorem]{Remark}

\begin{document}

\ifcolmsubmission
\linenumbers
\fi

\maketitle

\begin{abstract}
Discrete diffusion models have recently emerged as strong alternatives to autoregressive language models, matching their performance through large-scale training. However, inference-time control remains relatively underexplored. In this work, we study how to steer generation toward desired rewards without retraining the models. Prior methods typically resample or filter \emph{within a single denoising trajectory}, optimizing rewards step-by-step without trajectory-level refinement. We introduce particle Gibbs sampling for diffusion language models (PG-DLM), an inference-time algorithm enabling \emph{trajectory-level refinement}. PG-DLM constructs a Markov chain over full denoising trajectories and applies a conditional sequential Monte Carlo kernel to resample them.  By doing so, PG-DLM introduces a new scaling axis, the number of refinement iterations, which is unavailable to prior methods. 
Increasing iterations remains effective even as gains from adding more parallel samples saturate. Furthermore, PG-DLM enables adaptive compute allocation by performing additional iterations only when needed, leading to further efficiency gains.
We derive theoretical guarantees for convergence and variance bounds, and analyze trade-offs across different scaling axes. 
Empirically, PG-DLM outperforms prior methods across compute budgets on reward-guided generation tasks. On GSM8K, it achieves 90.07\% accuracy with 2.9 particles on average and 94.47\% accuracy with 24.8 particles on average.

\end{abstract}

\section{Introduction}
\label{sec:intro}
Recent advances in discrete diffusion models have established them as strong alternatives to autoregressive language models for text generation~\citep{austin2021d3pm,lou2023discrete,sahoo2024simple,shi2024simplified,zheng2024masked,nie2024scaling}. 
By scaling model size and training data, diffusion language models (DLMs) now match or surpass autoregressive LLMs on tasks like code generation and math reasoning, as demonstrated by models such as LLaDA-8B~\citep{nie2025large} and Dream-7B~\citep{ye2025dream}.

While this progress has focused primarily on \emph{training-time scaling}, which quickly becomes computationally expensive, a complementary and more efficient strategy remains underexplored: steering DLMs at \emph{inference time} toward desired attributes without modifying the underlying model.
Examples include generating texts with controlled toxicity~\citep{Dathathri2020Plug,keskar2019ctrl} or improved reasoning accuracy~\citep{snell2024scaling}.
This is typically formalized as sampling from a reward-weighted posterior:
$
p^*(\rvx_0\given\rvc) \propto p_\theta(\rvx_0\given\rvc)\exp\left(r(\rvc, \rvx_0)/\beta\right),
$
where $p_\theta(\rvx_0 \given \rvc)$ is the pretrained DLM, $r(\rvc, \rvx_0)$ is a reward function scoring the output $\rvx_0$ given prompt $\rvc$, and $\beta > 0$ controls reward strength~\citep{rafailov2024direct,korbak2022rl}.

To sample from the reward-weighted posterior at inference time, prior work has explored search-based strategies~\citep{ma2025inference} and particle-based methods like Best-of-$n$ and sequential Monte Carlo (SMC), including FK-Steering~\citep{singhal2025fksteering}, which scales the number of parallel samples. Another line uses predictor-corrector and remasking strategies~\citep{wang2025remasking,lezama2022discrete}, scaling via more denoising steps. {These methods maintain multiple parallel samples, each following \emph{a single denoising trajectory} $\rvx_T,\cdots,\rvx_0$, sampled step-by-step from $t\!=\!T$ to $t\!=\!0$, with resampling at intermediate timesteps}.
They do not perform \emph{trajectory-level refinement}, i.e., iteratively updating entire generations $\rvx_{0:T}$ across multiple passes.
\chg{More recent search-based methods~\citep{zhang2025inferencetimescalingdiffusionmodels,jain2025diffusiontreesamplingscalable} achieve trajectory-level refinement by revisiting full generations via backtracking in a search tree.} In contrast, we introduce a particle-based framework that performs trajectory-level refinement through iterative resampling of complete trajectories, enabling principled probabilistic inference and adaptive compute allocation.
\begin{figure*}[t]
    \centering
    \includegraphics[width=0.9\textwidth]{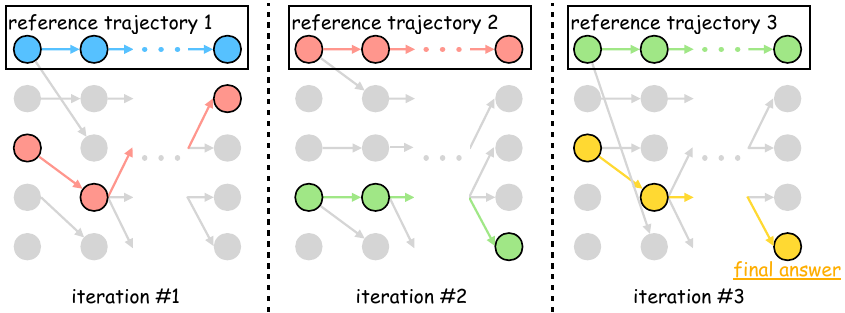}
    \caption{Illustration of \method{}. At each iteration, a reference trajectory is fixed (top row); new trajectories are generated and resampled (gray). We sample the next reference trajectory according to the final weights (colored), enabling iterative refinement. The final outputs are selected after multiple iterations.}
    \label{fig:intro}
\end{figure*}

In this paper, we introduce \textbf{particle Gibbs sampling for diffusion language models (PG-DLM)}. Built on the particle Gibbs framework~\citep{andrieu2010particle}, PG-DLM runs multiple full generation passes over a sequence of iterations.\footnote{Here, we refer to ``iteration'' as a \emph{trajectory-level update} ($m$ iterations) and ``timestep'' as the denoising steps within a single trajectory ($t=T,\dots,0$).} As shown in Figure~\ref{fig:intro}, in each iteration, it fixes the best trajectory from the previous iteration as a \emph{reference} and resamples the remaining trajectories via a conditional SMC kernel. This process introduces a new scaling axis, the number of refinement iterations, that is unavailable to prior single-pass methods.

We further investigate efficient allocation of inference-time compute within PG-DLM, analyzing trade-offs across four axes: particle Gibbs iterations, samples per iteration, denoising steps, and reward estimation cost. We find that scaling the number of iterations remains effective even as gains from increasing parallel samples saturate. Furthermore, PG-DLM enables adaptive compute allocation, performing additional iterations only when needed, leading to further efficiency gains. 
On GSM8K, PG-DLM (fixed) consistently outperforms SMC by 0.6\% to 0.9\% accuracy within the same compute budgets. With adaptive compute allocation, PG-DLM (adapt), the margin reaches 1.1\% to 2.8\%, achieving 90.07\% accuracy with an effective 2.9 particles on average and 94.47\% accuracy with 24.8 particles on average.

Our contributions are: \textbf{(1)} PG-DLM, a trajectory-level inference-time sampler for DLMs with convergence guarantees; \textbf{(2)} an adaptive compute allocation strategy that concentrates compute on harder instances (Section~\ref{sec:method}); and \textbf{(3)} empirical results showing it outperforms baselines across tasks, with scaling iterations improving reward  (Section~\ref{sec:exp}).

\section{Related Work}
\label{sec:related}

\paragraph{Inference-time scaling in LLMs.} Inference-time scaling has been extensively studied in autoregressive LLMs, where boosting compute during generation often proves more efficient than training-time scaling~\citep{snell2024scaling}. Techniques like beam search, diverse verifier trees~\citep{beeching2024scaling}, and particle filtering~\citep{puri2025probabilistic,lew2023sequential} have enhanced mathematical reasoning and constrained generation. SMC methods have recently been applied to autoregressive LLMs for controllable generation and probabilistic inference: \citet{loula2025syntactic} uses SMC for syntactic and semantic control, while \citet{zhao2024probabilistic} develops twisted SMC for probabilistic inference in language models.
While LLMs benefit from these tools, analogous strategies for discrete diffusion models remain underdeveloped.

\paragraph{Inference-time scaling in diffusion models.} A core approach to scaling diffusion inference is increasing denoising steps: \citet{ma2025inference} explore search-based strategies, while \citet{wang2025remasking} dynamically extend trajectories via re-masking in masked models. For search-based methods, 
\citet{zhang2025inferencetimescalingdiffusionmodels,jain2025diffusiontreesamplingscalable} incorporate mechanisms that can revisit full generation via backtracking in the search tree for trajectory-level refinement, while \cite{guo2025trainingfreeguidancedifferentiabilityscalable} performs tree search without explicit refinement of full generations. Particle-based methods scale parallel samples to guide toward high-reward regions~\citep{singhal2025fksteering,kim2025inference}, while reinforcement learning optimizes reasoning in diffusion LLMs~\citep{zhao2025d1}. SVDD~\citep{li2024derivative} proposes soft value-based decoding that selects samples at each timestep based on learned value functions, avoiding the need for differentiable rewards.
Predictor-corrector schemes~\citep{lezama2022discrete,zhao2024informed,gat24discrete} and classifier guidance~\citep{schiff2024simple} further improve controllability and quality in discrete settings. In continuous diffusion, particles aid inverse problems~\citep{wu2023practical,dou2024diffusion,nazemi2024particle} and generation~\citep{kim2025inference}.

\paragraph{Gibbs sampling in diffusion models.}
Gibbs sampling has been explored in the diffusion literature for iterative refinement of generated samples. \citet{bengio2013generalized} established an early connection between denoising auto-encoders and Markov chains whose stationary distribution matches the data distribution. More recently, \citet{zhang2023moment} and \citet{chen2024diffusive} proposed methods that alternate between forward corruption and reverse denoising for iterative refinement. These methods operate by iteratively refining individual samples through repeated denoising. In contrast, our work applies particle Gibbs~\citep{andrieu2010particle} to iteratively refine entire denoising trajectories: a reference trajectory is held fixed while the remaining particles are resampled, targeting a reward-weighted distribution.

\section{Background}
\label{sec:background}

\subsection{Discrete Diffusion Language Models}
Discrete diffusion language models (DLMs)~\citep{austin2021d3pm,lou2023discrete,shi2024simplified,sahoo2024simple} have emerged as a powerful alternative to autoregressive models, matching their performance through large-scale
training~\citep{nie2025large,ye2025dream}. Unlike continuous diffusion models~\citep{sohl2015deep,ddpm,song2019generative}, DLMs operate on discrete token spaces, reversing a masking corruption process to iteratively denoise sequences.

Let $\rvx_0 = (x_1, \dots, x_L)$ denote a clean sequence of $L$ tokens, where each token $x_i \in \mathcal{X}$ is a one-hot vector; $\rvx_t$ the corrupted state at time $t \in [0, T]$; and $\rvm$ the [MASK] token. 
The forward process $q$ gradually corrupts $\rvx_0$ by replacing tokens with $\rvm$:
\begin{equation}
    q(\rvx_t \mid \rvx_0) = \mathrm{Cat}(\rvx_t; \alpha_t \rvx_0 + (1 - \alpha_t) \rvm),
    \label{eq:forward-masked}
\end{equation} 
where $\mathrm{Cat}(\cdot)$ denotes the categorical distribution over the vocabulary, and the noise schedule $\alpha_t$ decreases monotonically from $\alpha_0 = 1$ to $\alpha_T = 0$. This enables a closed-form posterior:
\begin{equation}
q(\rvx_{t-1}\mid\rvx_t,\rvx_0)=
\begin{cases}
\mathrm{Cat}(\rvx_{t-1};\,\rvx_t)\;\;(\rvx_t\neq\rvm),\\
\mathrm{Cat}(\rvx_{t-1};\,\pi_t)\;\;(\rvx_t=\rvm)
\end{cases}
\label{eq:backward}
\end{equation}
where $\pi_t:=\frac{\alpha_{t-1}-\alpha_t}{1-\alpha_t}\rvx_0
+\frac{1-\alpha_{t-1}}{1-\alpha_t}\rvm.$
To approximate this posterior, DLMs train a denoising model $\rvx_\theta(\rvx_t)$  that outputs a distribution over the vocabulary $\mathcal{X}$ to predict $\rvx_0$ from $\rvx_t$. 
The resulting backward transition is $p_\theta(\rvx_{t-1} \mid \rvx_t) = q(\rvx_{t-1} \mid \rvx_t, \rvx_\theta(\rvx_t))$. 
The model is trained by minimizing the negative evidence lower bound (NELBO) to maximize data likelihood:
\begin{equation*}
    -\log p_\theta(\rvx_0)\leq  \sum_{t=1}^{T} \mathbb{E}_{q(\rvx_t \mid \rvx_0)} \left[
    \frac{\alpha_t - \alpha_{t-1}}{1 - \alpha_t} \log \left( \rvx_\theta(\rvx_t)^\top \rvx_0 \right)
    \right].
    \label{eq:loss-masked}
\end{equation*}

\subsection{Reward-Weighted Generation of Diffusion Language Models}
\label{sec:background-reward}

In this work, we align  diffusion language models $p_\theta(\rvx_0 \mid \rvc)$ with task-specific rewards $r(\rvc, \rvx_0)$, where $\rvc$ is a prompting prefix and $\rvx_0$ the generated sequence. Examples include generating high-quality text or sentiment control~\citep{Dathathri2020Plug,keskar2019ctrl}.
Following~\citet{jaques2017sequence,ouyang2022training}, this can be formalized as a KL-regularized reinforcement learning objective, where we maximize expected reward while remaining close to the base model $p_\theta$ to get optimal target distribution $p^*$:
\begin{align}
\label{eq:rl}
p^*(\rvx_0 \given \rvc)
= \arg\max_{p}\;
& \mathbb{E}_{\rvx_0 \sim p}\!\left[r(\rvc,\rvx_0)\right] - \beta\,\mathrm{KL}\!\left(
p(\rvx_0 \given \rvc)\,\|\,p_\theta(\rvx_0 \given \rvc)
\right).
\end{align}
where hyperparameter $\beta \!>\! 0$ controls the trade-off between reward maximization and divergence from base model. This objective has a closed-form solution~\citep{rafailov2024direct}
\begin{equation}
    p^*(\rvx_0 \given \rvc) \propto p_\theta(\rvx_0 \given \rvc) \cdot \exp\left(r(\rvc, \rvx_0)/\beta\right),
    \label{eq:p-target}
\end{equation}
which reweights the base model distribution toward higher-reward generations.
While fine-tuning methods can align base models $p_\theta$ to the target $p^*$~\citep{clark2023directly,black2024ddpo,fan2024dpok,wallace2024diffusion}, we instead pursue \emph{inference-time} approximation via sampling. 

\section{Method}
\label{sec:method}
In this section, we first derive the reward-weighted denoising conditionals from an RL perspective and present sequential Monte Carlo (SMC) as a baseline sampler. We then introduce particle Gibbs sampling for diffusion language models (PG-DLM), a trajectory-level refinement method that overcomes SMC's limitations, and demonstrate its generality while proving convergence guarantees.

\subsection{Problem Setup and SMC for Diffusion Models}
In the backward process of a DLM $p_\theta(\rvx_0 \given \rvc)$, generation begins with a fully masked sequence $\rvx_T = \rvm$ and iteratively unmasks tokens via the denoising model $p_\theta(\rvx_{t-1} \given \rvc, \rvx_t)$, yielding a full \emph{denoising trajectory} $\rvx_{0:T} = \{\rvx_0, \dots, \rvx_T\}$.
However, to sample from the reward-weighted target distribution $p^*(\rvx_0 \given \rvc)$ as in Equation~\ref{eq:p-target}, one must use the corresponding conditional distributions $p^*(\rvx_{t-1} \given \rvc, \rvx_t)$ at each timestep. Building on prior work in the continuous setting~\citep{uehara2024fine,uehara2024bridging}, we present the tractable formulation in the discrete setting:
\begin{multline}
p^*(\rvx_{t-1}\given\rvc,\rvx_t)
\propto p_\theta(\rvx_{t-1}\given\rvc,\rvx_t)
\cdot \exp\!\left(\left(r(\rvc,\rvx_{t-1})-r(\rvc,\rvx_t)\right)/\beta\right),
\\
r(\rvc,\rvx_t)
= \beta \log \mathbb{E}_{p_\theta(\rvx_0\given\rvc,\rvx_t)}
\!\left[\exp\!\left(r(\rvc,\rvx_0)/\beta\right)\right].
\label{eq:p_cond}
\end{multline}
Here, $r(\rvc, \rvx_t)$ defines a \emph{partial reward function} for the noisy intermediate state $\rvx_t$, representing the expected future reward at timestep $t$ under the pretrained model $p_\theta$.
This formulation shows that the conditional $p^*(\rvx_{t-1} \given \rvc, \rvx_t)$ is a reward-weighted posterior, with weights given by the difference in partial rewards. It mirrors the reward-weighted objective in Equation~\ref{eq:p-target} through timestep-wise decomposition, incorporating the reward difference at each step. \chg{While we formally derive the reward-difference structure from an RL perspective, where the difference in rewards across timesteps $r(\rvc, \rvx_{t-1}) - r(\rvc, \rvx_t)$ is used to guide generation, similar formulations have been used as sampling heuristics in prior work~\citep{singhal2025fksteering,wu2023practical} without establishing explicit connections to RL objectives.} This grounding not only justifies the partial-reward weighting but also enables extensions to other KL-regularized tasks.

Given the reward-weighted conditional distribution $p^*(\rvx_{t-1} \given \rvc, \rvx_t)$ as in Equation~\ref{eq:p_cond}, one intuitive way to generate samples from this target is to first draw samples from the base model $p_\theta(\rvx_{t-1} \given \rvc, \rvx_t)$ and then resample them based on their reward weights.
This backward process, iterated from $t\!=\!T$ down to $t = 0$, is known as \emph{sequential Monte Carlo (SMC)} or \emph{particle filtering}, where $p_\theta$ is the \emph{proposal distribution} and $p^*$ the \emph{target distribution}~\citep{naesseth2019elements,doucet2001introduction}.

Concretely, the SMC algorithm proceeds as follows: At timestep $T$, we initialize $k$ samples as masked sequences $\rvx_{T}^{i} = \rvm$ for $i = 1, \dots, k$. Then, for each timestep $t$, the process involves:
(1) \textbf{proposing} $\bar{\rvx}_{t-1}$ samples from the proposal distribution $p_\theta(\rvx_{t-1} \given \rvc, \rvx_t)$ for each $\rvx_t$;
(2) \textbf{reweighting} by computing importance weights $w_{t-1} = \exp\left(\left(r(\rvc, \bar{\rvx}_{t-1}) - r(\rvc,\bar{\rvx}_t)\right)/\beta \right)$ as in Equation~\ref{eq:p_cond};
and (3) \textbf{resampling} with replacement from $\bar{\rvx}_{t-1}$ according to the normalized weights $w_{t-1}$ to form $\rvx_{t-1}$.
This method has been referred to as Feynman-Kac Steering~\citep{singhal2025fksteering} in the context of reward-weighted generation for diffusion models.

\subsection{A Particle Gibbs Sampler}

While SMC provides a simple way to scale inference-time compute by increasing the number of samples, it has several limitations that hinder effective reward alignment in DLMs.
\chg{Samples evolve as parallel trajectories interacting only via reweighting and resampling, limiting inter-sample correlations between them.} Moreover, it performs a ``one-shot'' approximation in a single backward pass from $t=T$ to $t=0$ without iterative \emph{trajectory-level refinement}. Finally, SMC is prone to weight degeneracy and high variance in importance weights under skewed reward landscapes~\citep{naesseth2019elements}.

\begin{algorithm}[t]
\caption{Particle Gibbs for Diffusion Language Models}
\label{alg:pg}
\footnotesize
\algnewcommand\algorithmicinput{\textbf{Input: }}
\algnewcommand\Input{\State \algorithmicinput}
\algnewcommand\algorithmicoutput{\textbf{Output: }}
\algnewcommand\Output{\State \algorithmicoutput}
\begin{algorithmic}[1]
\Input iterations $m$, particle count $k$, timesteps $T$,
partial reward samples $\phi$,
\par\hspace{\widthof{{In}}}
hyperparameter $\beta$, reward model $r(\rvc, \rvx_0)$,
diffusion model $p_\theta(\rvx_{t-1} \given \rvc, \rvx_t)$
\Output sample from $p^*(\rvx_0 \given \rvc) \!\propto\! p_\theta(\rvx_0 \given \rvc) \exp\left(r(\rvc, \rvx_0)/\beta\right)$
\State Sample initial reference trajectory $\rvx'_{0:T} \sim p_\theta(\rvx_0 \given \rvc)$ via backward process
\For{$\text{iter} = 1$ to $m$}
  \State Initialize $k$ samples $\rvx_T^{(i)} = \rvm$ for $i = 1, \dots, k$
  \For{$t = T$ to $1$}
    \State Fix reference $\bar{\rvx}_{t-1}^{(k)} = \rvx'_{t-1}$
    \State Propose $\bar{\rvx}_{t-1}^{(i)} \sim p_\theta(\rvx_{t-1} \given \rvc, \rvx_t^{(i)})$ for $i = 1, \dots, k-1$
    \State Estimate reward
    $\hat{r}(\rvc, \bar{\rvx}_{t-1}^{(i)}) = \beta \log \left( \frac{1}{\phi} \sum_{j=1}^\phi \exp\left( r(\rvc, \tilde{\rvx}_0^{(j)})/\beta \right) \right)$
    where $\tilde{\rvx}_0^{(j)} \sim p_\theta(\rvx_0 \given \rvc, \bar{\rvx}_{t-1}^{(i)})$
    \State Compute importance weights for $i = 1, \dots, k$: 
    $\bar{w}_{t-1}^{(i)} = \exp\left( \left(\hat{r}(\rvc, \bar{\rvx}_{t-1}^{(i)}) - \hat{r}(\rvc, \rvx_t^{(i)})\right)/\beta \right)$
    \State Normalize $w_{t-1}^{(i)} = \bar{w}_{t-1}^{(i)} / \sum_{j=1}^k \bar{w}_{t-1}^{(j)}$ for $i = 1, \dots, k$
    \State Sample with replacement $\rvx_{t-1}^{(i)} \sim \{\bar{\rvx}_{t-1}^{(j)}, w_{t-1}^{(j)}\}_{j=1}^k$ for $i = 1, \dots, k-1$
    \State Fix $\rvx_{t-1}^{(k)} = \rvx'_{t-1}$
  \EndFor
  \State Compute unnormalized final weights for $i = 1, \dots, k$:
  $\bar{w}_0^{(i)} = \exp\left( r(\rvc, \rvx_0^{(i)})/\beta \right)$
  \State Normalize $w_0^{(i)} = \bar{w}_0^{(i)} / \sum_{j=1}^k \bar{w}_0^{(j)}$ for $i = 1, \dots, k$
\State Sample $i^* \sim \text{Cat}(w_0^{(1)}, \dots, w_0^{(k)})$ and update reference $\rvx'_{0:T} \leftarrow \rvx_{0:T}^{(i^*)}$
\EndFor
\State \Return sample $\rvx'_0$ or weighted samples $\{\rvx_0^{(i)}, w_0^{(i)}\}_{i=1}^k$
\end{algorithmic}
\end{algorithm}
To address these limitations, we propose an iterative trajectory-level sampling framework called \textbf{particle Gibbs for diffusion language models (\method{})}. Intuitively, as shown in Figure~\ref{fig:intro}, \method{} refines high-reward trajectories across multiple sequential denoising processes: we begin by generating a batch of candidate trajectories $\rvx_{0:T}$, sample a trajectory according to the weights as the \emph{reference trajectory}, and then resample new trajectories guided by this reference, exploring variations around it. This process is repeated iteratively, correlating samples across multiple denoising passes and leveraging the full capacity of $p_\theta$. As shown later, this yields better reward optimization while maintaining generation quality.

Formally, \method{} is a particle Gibbs sampler~\citep{andrieu2010particle}, a Markov Chain Monte Carlo (MCMC) algorithm that iteratively refines complete trajectories $\rvx_{0:T}$. It uses a \emph{conditional sequential Monte Carlo (SMC)} transition kernel to update the trajectories.
As detailed in Algorithm~\ref{alg:pg}, \method{} begins by generating one sample from the base model as an initial reference trajectory (line 3), then performs $m$ iterations of conditional SMC updates (lines 4--18). 
In each iteration, the conditional SMC update proceeds backward through each timestep $t$ by: 
(1) \textbf{fixing} the reference trajectory deterministically as the $k$-th sample (line 7); 
(2) \textbf{proposing} $k-1$ new samples from the base model (line 8); 
(3) \textbf{reweighting} all $k$ samples, including the fixed $k$-th one (lines 9-11); and 
(4) \textbf{resampling} the first $k-1$ candidates with replacement, proportional to their normalized weights, while keeping the $k$-th sample fixed (lines 12-13). 
After each iteration, the new reference trajectory is sampled from the current batch (lines 15-17). This iterative process allows the final trajectory to closely approximate the target distribution $p^*(\rvx_0 \given \rvc)$.

\paragraph{Adaptive Compute Allocation.}
PG-DLM naturally supports adaptive compute allocation by performing 
additional iterations only when needed. Starting from an initial 
trajectory, we check whether the reward of the final output exceeds 
a threshold; if so, we return early without further iterations. 
This allocates more compute to harder instances while avoiding 
unnecessary work on easier ones. We refer to this algorithm as \method{}~(adapt).

\paragraph{Generality Across Backward Processes. }
The \method{} framework is broadly compatible with arbitrary backward transitions $p(\rvx_{t-1} \given \rvc, \rvx_t)$ in DLMs. Examples include the standard unmasking in MDLM~\citep{sahoo2024simple} (Equation~\ref{eq:backward}), greedy low-entropy unmasking in LLaDA~\citep{nie2025large}, and correction/re-masking mechanisms~\citep{wang2025remasking, lezama2022discrete}. 
\subsection{Convergence and Variance Analysis}
For \method{}, convergence depends on accurately computing the importance weights. As shown in Algorithm~\ref{alg:pg}, we approximate the partial reward using $\phi$ Monte Carlo samples $\rvx_0 \sim p_\theta(\rvx_0 \given \rvc, \rvx_t)$.

\begin{remark}
\label{rem:partial_reward}
Let \( p^*(\rvx_0 \mid \rvc) \propto p_\theta(\rvx_0 \mid \rvc) \cdot \exp\left(r(\rvc, \rvx_0)/\beta\right) \) be the target distribution, where \( p_\theta(\rvx_0 \mid \rvc) \) is a discrete diffusion model with \( T \) denoising steps. The samples $\rvx_0^{(j)} \sim p_\theta(\rvx_0 \given \rvc, \rvx_t)$ obtained via the $t$-step denoising process are i.i.d. By the strong law of large numbers, the partial reward estimator $\hat{r}(\rvc, \rvx_t) = \beta \log \frac{1}{\phi}\sum_{j=1}^{\phi}\left[ \exp\left( r(\rvc, \rvx_0^{(j)})/\beta \right) \right]$ (cf. Equation~\ref{eq:p_cond}) converges to the true partial reward $r(\rvc, \rvx_t)$ as \( \phi \to \infty \).
\end{remark}
The reference trajectory in \method{} ensures that the conditional SMC updates leave the target distribution \emph{invariant} and \emph{ergodic} for $k \geq 2$~\citep{andrieu2010particle}. Under the assumptions that (i) partial rewards are estimated exactly ($\phi \to \infty$), and (ii) $k \geq 2$, combined with Remark~\ref{rem:partial_reward}, this directly yields Theorem~\ref{theorem1} on asymptotic consistency (adapted from~\citet{andrieu2010particle}) and Theorem~\ref{the:variance} on per-iteration variance bounds (adapted from~\citet{andrieu2010particle,chatterjee2018sample}).

\begin{theorem}[Asymptotic Consistency]
\label{theorem1}
Given Remark~\ref{rem:partial_reward}, the empirical distribution produced by \method{} converges almost surely to the target \( p^*(\rvx_0 \!\mid\!\rvc) \) as \( m \!\to\!\infty \), \( \phi\!\to\!\infty \), given \( k \!\geq\! 2 \). 
\end{theorem}

\begin{theorem}[Variance Bound]
\label{the:variance}
Given Remark~\ref{rem:partial_reward},
let the unnormalized target be $\tilde{p}(\rvx_{0:T}\!\given\!\rvc) \!=\! \gamma(\rvc, \rvx_0) \cdot p_\theta(\rvx_{0:T}\given \rvc)$,  where $\gamma(\rvc,\rvx_0)\!=\!\exp(r(\rvc, \rvx_0)/\beta)$. Its normalizing constant is $Z \!=\!\sum_{\rvx_{0:T}} \tilde{p}(\rvx_{0:T}\given \rvc)$. For the estimator $\widehat{Z}$ within a single iteration of \method{} with $k$ samples, the variance
\[
\mathrm{Var}(\widehat{Z}) \leq \frac{\mathrm{Var}_{p_\theta(\rvx_0 \given \rvc)}\left[\gamma(\rvc, \rvx_0)\right]}{k},
\]
where $\mathrm{Var}_{p_\theta(\rvx_0 \given \rvc)}\left[\gamma(\rvc, \rvx_0)\right] = \mathbb{E}_{p_\theta(\rvx_0 \given \rvc)}[\gamma(\rvc, \rvx_0)^2] - Z^2$.
\end{theorem}

As the number of iterations $m$ increases, the Markov chain over trajectories converges to the target distribution (Theorem~\ref{theorem1}), and we empirically observe reduced weight variance with additional iterations (Table~\ref{tab:ess}).
This variance bound shows that \method{}'s variance is determined by that of the reweighting function $\gamma(\rvc, \rvx_0) = \exp(r(\rvc, \rvx_0)/\beta)$ under the proposal $p_\theta(\rvx_0 \mid \rvc)$. For example, if $r(\rvc, \rvx_0)$ is constant, the proposal matches the target and $\mathrm{Var}(\widehat{Z}) = 0$; if $r(\rvc, \rvx_0)$ is highly peaked, $\gamma(\rvc, \rvx_0)$ has large variance, as the proposal fails to cover high-reward regions effectively, leading to inefficient sampling.

\section{Experiments}
\label{sec:exp}

\subsection{Baselines}
We compare \method{}~(fixed), which uses a fixed number of particle Gibbs iterations $m$, and \method{}~(adapt), which adaptively performs additional iterations when the reward falls below a threshold (cf.~Section~\ref{sec:method}), against the following inference-time baselines:
\begin{itemize}[leftmargin=*,nosep]
  \item \textbf{Greedy Decoding.} Deterministically unmasks the highest-probability position at each step. It uses a single trajectory without resampling, i.e., $n=1$.
  \item \textbf{Best-of-$n$ Sampling.} Generates $n$ independent trajectories from the base model and selects the one with the highest reward.
  \item \textbf{FK-Steering/SMC~\citep{singhal2025fksteering}.} A sequential Monte Carlo method that maintains $k$ parallel trajectories, with reweighting and resampling at intermediate timesteps based on estimated rewards.
\end{itemize}
Search-based methods are complementary; we focus on particle-based methods for controlled comparison under matched compute budgets.

We use $n$ to denote the total number of particles as a proxy for compute budget, e.g., number of function evaluations (NFEs), since reward evaluation incurs comparable overhead across methods. For \method{}~(fixed), $n = mk$, while for \method{}~(adapt), $n$ is determined dynamically. Greedy decoding has $n=1$, and for Best-of-$n$ and FK, $n$ is given directly.
\begin{figure}[t]
\centering

\begin{minipage}[t]{0.42\columnwidth}
  \vspace{-15pt}
  \centering
  \includegraphics[width=\columnwidth]{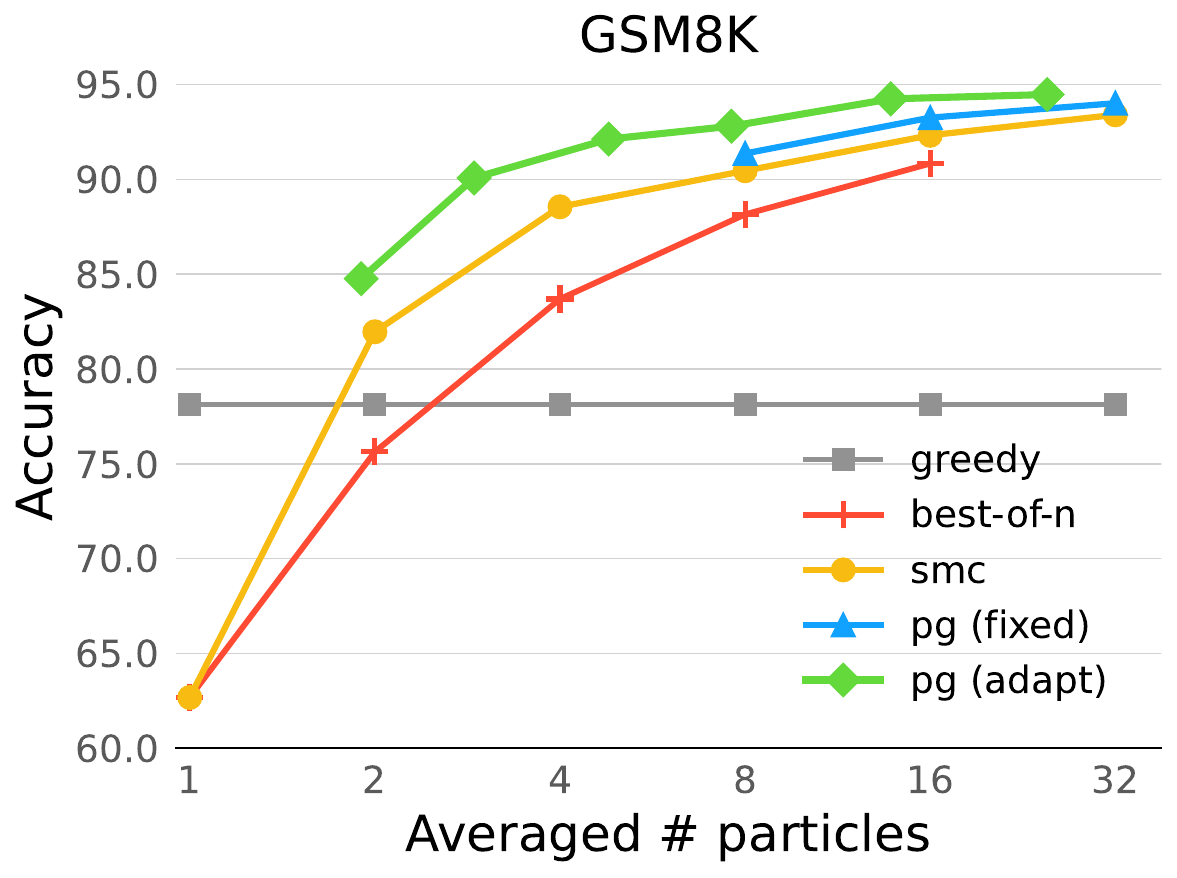}
  \label{fig:gsm8k_curve}
\end{minipage}
\hfill
\begin{minipage}[t]{0.56\columnwidth}
  \vspace{0pt}
  \centering
  \footnotesize
\setlength{\tabcolsep}{3pt}
\begin{tabular}{lccccc}
\toprule
Method & 2 & 4 & 8 & 16 & 32 \\
\midrule
SMC       & 81.96 & 88.55 & 90.45 & 92.32 & 93.40 \\
\method{}~(fixed)        & --    & --    & 91.36 & 93.25 & 94.01 \\
\method{}~(adapt)  &
\makecell{84.76 \\ \scriptsize (1.9)} &
\makecell{90.07 \\ \scriptsize (2.9)} &
\makecell{92.80 \\ \scriptsize (7.6)} &
\makecell{94.24 \\ \scriptsize (13.8)} &
\makecell{94.47 \\ \scriptsize (24.8)} \\
\midrule
$\Delta$(\method{}~(fixed)) & -- & -- & \textcolor{darkblue}{+0.91} & \textcolor{darkblue}{+0.93} & \textcolor{darkblue}{+0.61} \\
$\Delta$(\method{}~(adapt))  & \textcolor{darkgreen}{\textbf{+2.80}} & \textcolor{darkgreen}{\textbf{+1.52}} & \textcolor{darkgreen}{\textbf{+2.35}} & \textcolor{darkgreen}{\textbf{+1.92}} & \textcolor{darkgreen}{\textbf{+1.07}} \\
\bottomrule
\end{tabular}
  \label{tab:gsm8k}
\end{minipage}

\caption{\footnotesize{
Accuracy vs.\ compute (average number of particles) on GSM8K (left) with corresponding values (right).
For \method{}~(adapt), the effective number of particles is shown in parentheses.
Improvements ($\Delta$) are computed relative to SMC; 
\textcolor{darkgreen}{green} and \textcolor{darkblue}{blue} denote the best and second-best gains. Greedy and Best-of-$n$ appear in the plot but are omitted from the table for space.}
}
\label{fig:gsm8k_combined}
\end{figure}
\subsection{\chg{Math Reasoning Tasks}}
\label{sec:gsmk}

\myparagraph{Task and Dataset.} 
We evaluate our approach on GSM8K~\citep{cobbe2021gsm8k}, a standard benchmark for multi-step arithmetic reasoning. We use LLaDA-8B-Instruct~\citep{nie2025large} as the base model, and Qwen2.5-Math-PRM-7B~\citep{prmlessons} as the reward model. A key advantage of this reward model is its ability to compute rewards directly on prefixes, which is well-suited for estimating reward in blocked diffusion and avoids the need to sample full completions.

\myparagraph{Setup.} 
For all methods, we fix the maximum generation length to $L=512$, use $T=256$ denoising steps, and adopt a block size of 32. We vary the compute budget by controlling the total number of particles $n \in \{1,2,4,8,16,32\}$. For both SMC, which we reimplement, and \method{}, we perform resampling at the end of each block.
For \method{}~(fixed), we set $k=8$ and vary the number of particle Gibbs iterations $m \in \{1,2,4\}$, resulting in $n = mk$. For \method{}~(adapt), we start from an initial greedy decoding trajectory rather than a sampled one (cf. Algorithm 1), and perform additional particle Gibbs iterations only when the reward of the final output $\rvx_0$ is below a fixed threshold for early stopping. We vary $k$ across runs to match different compute budgets; as a result, the total number of particles $n$ is determined dynamically at runtime.

\myparagraph{Results.}
As shown in Figure~\ref{fig:gsm8k_combined}, \method{}~(fixed) consistently outperforms SMC by 0.6\% to 0.9\% accuracy within the same compute budgets. With adaptive compute allocation, the margin reaches 1.1\% to 2.8\%, with \method{}~(adapt) achieving 90.07\% accuracy with effective 2.9 particles on average and 94.47\% accuracy with effective 24.8 particles. These results highlight the benefit of trajectory-level refinement over locally improving individual decoding steps.

\subsection{Classifier-Based Reward}

\myparagraph{Task and Dataset.}
We evaluate three classifier-based reward functions for controllable generation: 
(1) \textit{Linguistic acceptability}, which favors grammatically correct sentences~\citep{warstadt2019neural}; 
(2) \textit{Toxicity control}, which identifies harmful content~\citep{logacheva2022paradetox}; and 
(3) \textit{Sentiment control}, which steers generation toward target sentiments~\citep{barbieri2020tweeteval}. 
We conduct experiments on two base models: MDLM~\citep{sahoo2024simple} and LLaDA-8B-Base~\citep{nie2025large}.

\begin{table*}[t]
\centering
\caption{\footnotesize{Controlled generation accuracy on LLaDA-8B-Base. Columns denote the total number of particles $n$. We compare \method{} with Best-of-$n$ and FK-Steering under matched compute budgets across reward functions.}}
\label{tab:control-gen-llada}
\scriptsize{
\setlength{\tabcolsep}{2.5pt} %
\begin{tabular}
{{L{68pt}
>{\columncolor{gray!20}}P{31pt}
>{\columncolor{gray!20}}P{31pt}
>{\columncolor{gray!20}}P{31pt}
P{31pt}P{31pt}P{31pt}
>{\columncolor{gray!20}}P{31pt}
>{\columncolor{gray!20}}P{31pt}
>{\columncolor{gray!20}}P{31pt}
}}
\toprule
\multirow{2}{*}{Method} & \multicolumn{3}{c}{CoLA $\uparrow$} & \multicolumn{3}{c}{Toxicity $\uparrow$} & \multicolumn{3}{c}{Sentiment $\uparrow$} \\
\cmidrule(lr){2-4} \cmidrule(lr){5-7} \cmidrule(lr){8-10}
& 4 & 16 & 64 & 4 & 16 & 64 & 4 & 16 & 64 \\
\midrule
{Best-of-$n$} &
74.2 $\pm$ 2.9 & 88.8 $\pm$ 2.3& 87.7 $\pm$ 0.9 &
 2.4 $\pm$ 0.2 & 9.0 $\pm$ 3.8& 29.2 $\pm$ 3.7 &
48.2 $\pm$ 2.9 & 85.7 $\pm$ 0.6 & 98.1 $\pm$ 1.2\\

FK-Steering &
74.1 $\pm$ 1.3  &        87.9  $\pm$  1.5  &        88.2 $\pm$ 2.4  &

\textbf{9.0}  $\pm$ 1.5  &        43.2 $\pm$ 2.7  &        80.9  $\pm$ 1.4 &

\textbf{69.4} $\pm$ 1.2 &        96.0  $\pm$ 1.2 &        \textbf{99.7}   $\pm$ 0.3\\

{\method{}}  &
\textbf{{77.8}} $\pm$ 2.2   &        \textbf{91.1} $\pm$ 3.1 &        \textbf{90.6} $\pm$ 0.2  &

      {8.3} $\pm$ 1.8 &        \textbf{48.3} $\pm$ 1.5  &        \textbf{89.1} $\pm$ 2.3   &

      66.6  $\pm$ 1.0 &       \textbf{96.4} $\pm$ 1.1  &        \textbf{99.7} $\pm$  0.2 \\
\bottomrule
\end{tabular}
}
\end{table*}
\begin{table*}[t]

\centering
\caption{\footnotesize{
Controlled generation accuracy on MDLM. Columns denote total number of particles $n$, and rows vary reward estimation (Beam/Random), backward process (MDLM/ReMDM). We compare \method{} with Best-of-$n$ and FK-Steering under matched compute budgets across reward functions.
}}
\label{tab:control-gen-mdlm}
\scriptsize
\setlength{\tabcolsep}{2.5pt} %
\begin{tabular}{{L{68pt}
>{\columncolor{gray!20}}P{31pt}
>{\columncolor{gray!20}}P{31pt}
>{\columncolor{gray!20}}P{31pt}
P{31pt}P{31pt}P{31pt}
>{\columncolor{gray!20}}P{31pt}
>{\columncolor{gray!20}}P{31pt}
>{\columncolor{gray!20}}P{31pt}
}}
\toprule
\multirow{2}{*}{Method} & \multicolumn{3}{c}{CoLA $\uparrow$} & \multicolumn{3}{c}{Toxicity $\uparrow$} & \multicolumn{3}{c}{Sentiment $\uparrow$} \\
\cmidrule(lr){2-4} \cmidrule(lr){5-7} \cmidrule(lr){8-10}
& 4 & 16 & 64 & 4 & 16 & 64 & 4 & 16 & 64 \\
\midrule
\multicolumn{1}{l}{Best-of-$n$}
& 71.3 $\pm$ 1.3 & 96.9 $\pm$ 1.6 & 95.8 $\pm$ 1.3
& 1.9 $\pm$ 0.4 & 11.4 $\pm$ 1.0 & 33.8 $\pm$ 2.8
& 36.7 $\pm$ 3.7 & 79.9 $\pm$ 1.0 & 99.6 $\pm$ 0.2 \\

\midrule
\multicolumn{1}{l}{FK-Steering} \\
\midrule

~~(Rand, MDLM) & 
 48.4 $\pm$   3.2      &   76.2 $\pm$   0.4       &        83.1 $\pm$   4.8       & 

  3.4 $\pm$   0.2       &        34.0 $\pm$   3.4       &        76.8 $\pm$   1.1       &

 33.6 $\pm$   3.7       &        89.2 $\pm$   1.5       &        98.9 $\pm$   0.5       \\
~~(Rand, ReMDM)&
 87.4 $\pm$   1.7       &        93.6 $\pm$   1.0       &        92.9 $\pm$   1.3       &

 16.9 $\pm$   0.7      &        89.7 $\pm$   1.3       &        97.6 $\pm$   0.2       &

 67.7 $\pm$   2.8       &        97.9 $\pm$   0.7       &        99.4 $\pm$   0.2       \\
~~(Beam, MDLM) &
 66.6 $\pm$   1.7      &        94.8 $\pm$   0.2       &        97.8 $\pm$   1.0       &

 11.2 $\pm$   1.1       &        81.9 $\pm$   3.0       &        96.8 $\pm$   1.0       &

 57.6 $\pm$   5.9       &        94.2 $\pm$   0.8       &        99.2 $\pm$   0.2       \\
~~(Beam, ReMDM) & 
 91.7 $\pm$   0.9       &        97.8 $\pm$   0.7       &        97.5 $\pm$   0.2       &

 24.6 $\pm$ 0.7       &        95.4 $\pm$   0.7       &        98.7 $\pm$   0.3       &

 72.3 $\pm$   4.3       &        96.1 $\pm$   1.1       &        99.2 $\pm$   0.2        \\

\midrule
\method{} \\
\midrule
~~(Rand, MDLM) & 
 29.8 $\pm$   3.1 &        80.0 $\pm$   1.2       &        89.4 $\pm$   1.1       &

1.3 $\pm$   0.0 &        26.8 $\pm$   2.7       &        75.1 $\pm$   2.7       &

 12.8 $\pm$   2.0 &        82.7 $\pm$   2.1       &        99.1 $\pm$   0.5       \\
~~(Rand, ReMDM) &
 74.8 $\pm$   3.0&         97.4 $\pm$   0.7       &        98.7 $\pm$   0.7       & 

  1.6 $\pm$   0.5      &        84.8 $\pm$   0.8       &        96.4 $\pm$   1.8       &

 24.7 $\pm$   1.2       &        96.0 $\pm$   0.9       &        99.6 $\pm$   0.5       \\
~~(Beam, MDLM)  &
 37.3 $\pm$   2.4       &        88.0 $\pm$   1.0       &        96.8 $\pm$   0.5       &

  1.3 $\pm$   0.5      &        78.8 $\pm$   2.0       &        97.2 $\pm$   1.2       &

 21.8 $\pm$   1.7      &        94.4 $\pm$   0.5       &        99.0 $\pm$   0.3       \\
~~(Beam, ReMDM) 
& {77.3} $\pm$   2.0 & {97.3} $\pm$ 0.9 & {99.1} $\pm$ 0.5
& {1.4} $\pm$ 0.7 & {91.1} $\pm$ 1.0 & {98.1} $\pm$ 1.1
& {23.8} $\pm$ 2.2 & {96.2} $\pm$ 1.3 & {99.1} $\pm$ 0.2 \\
\bottomrule
\end{tabular}
\end{table*}

\myparagraph{Setup and Ablation.}
Following prior work~\citep{singhal2025fksteering,han2023ssd}, we generate 300 sequences with maximum length $L=50$ across various compute budgets. For \method{}, we fix $m = 1$, as performance on these control tasks is already near saturation (often close to 100\% accuracy), leaving little headroom for further gains from iteration scaling. In contrast, as shown in Section~\ref{sec:gsmk}, increasing $m$ yields consistent improvements on more challenging reasoning tasks. For MDLM, we use $T=1024$ denoising steps, while for LLaDA we use $T=50$. We report mean accuracy over 3 random seeds. 

We further examine the effect of the backward process on MDLM by comparing the vanilla backward dynamics with the ReMDM variant~\citep{wang2025remasking}. In addition, we study two partial reward estimation strategies: \textbf{random sampling} and \textbf{beam sampling}.

\myparagraph{Results and Analysis.}
Tables~\ref{tab:control-gen-llada} and~\ref{tab:control-gen-mdlm} compare all methods under fixed compute budgets $n$.
With $m=1$, \method{}~(fixed) reduces to a single conditional-SMC pass with one particle reserved for the reference, leaving $k-1$ for exploration. This overhead dominates at small $n$, where FK-Steering performs better, and diminishes as the budget grows, with \method{} matching or exceeding FK-Steering at $n=64$ most of the time. These control tasks are near saturation at $m=1$ and do not exercise the iteration axis where PG-DLM's gains come from.

As shown in Table~\ref{tab:control-gen-mdlm}, ReMDM consistently achieves stronger performance, demonstrating both the generality of our approach across different backward processes and its ability to benefit from improved diffusion dynamics. 
Finally, beam sampling outperforms random sampling due to its lower variance in reward estimation.

\subsection{Scaling Analysis}
\myparagraph{Effective Sample Size to Measure Convergence.}
We assess the convergence of \method{} using the \emph{effective sample size (ESS)}, computed from normalized importance weights $w_i$ for $i=1,\dots,k$ at the final timestep of each iteration: $\text{ESS} = 1 / \sum_{i=1}^k w_i^2$. 
ESS reflects the weight concentration per iteration and ranges from 1 to $k$, with higher values indicating more uniform weights and lower variance. As shown in Table~\ref{tab:ess}, ESS approaches $k$ after a single iteration and continues to increase with more iterations, demonstrating efficient convergence and reduced weight degeneracy.

\begin{figure*}[h]
    \centering
    \footnotesize{
    \begin{subfigure}{0.24\linewidth}
        \centering
        \includegraphics[width=\linewidth]{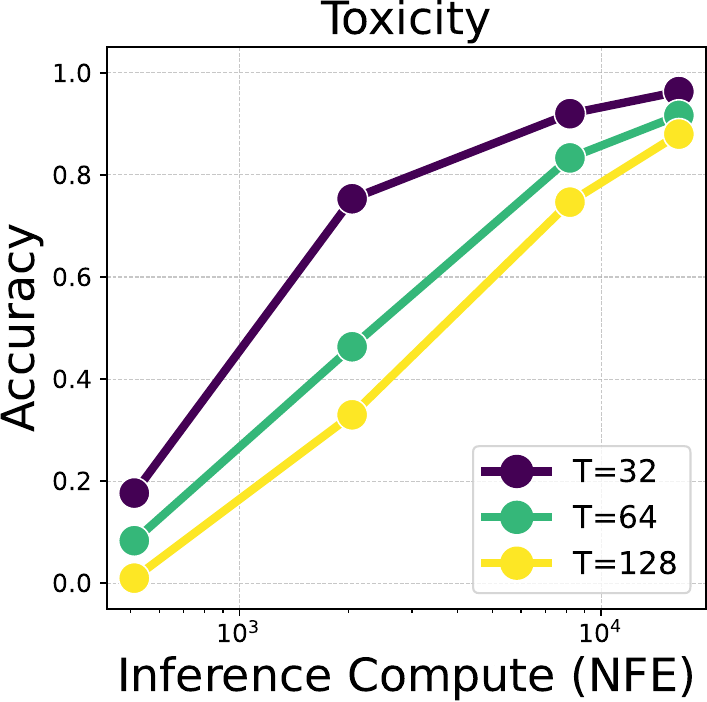}
        \caption{{LLaDA ($k$ via x-axis)}}
        \label{fig:llada-toxicity}
    \end{subfigure}
    \hfill
    \begin{subfigure}{0.24\linewidth}
        \centering
        \includegraphics[width=\linewidth]{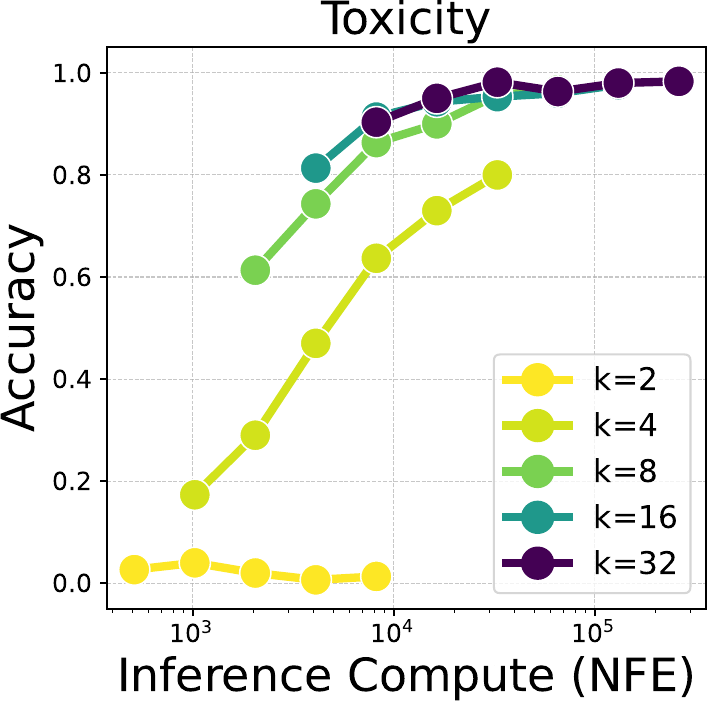}
        \caption{MDLM ($T$ via x-axis)}
        \label{fig:mdlm-toxicity}
    \end{subfigure}
    \hfill
    \begin{subfigure}{0.24\linewidth}
        \centering
        \includegraphics[width=\linewidth]{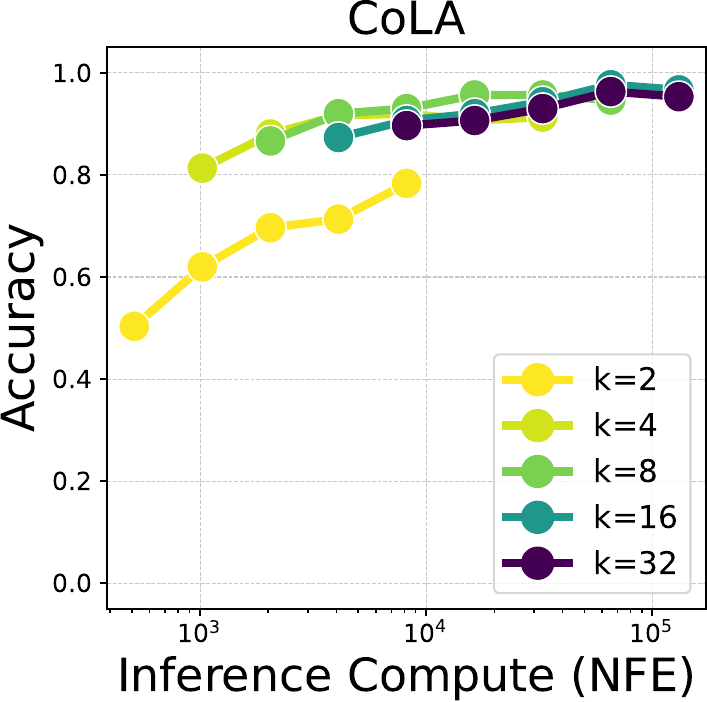}
        \caption{MDLM ($T$ via x-axis)}
        \label{fig:mdlm-cola}
    \end{subfigure}
    \hfill
    \begin{subfigure}{0.24\linewidth}
        \centering
        \includegraphics[width=\linewidth]{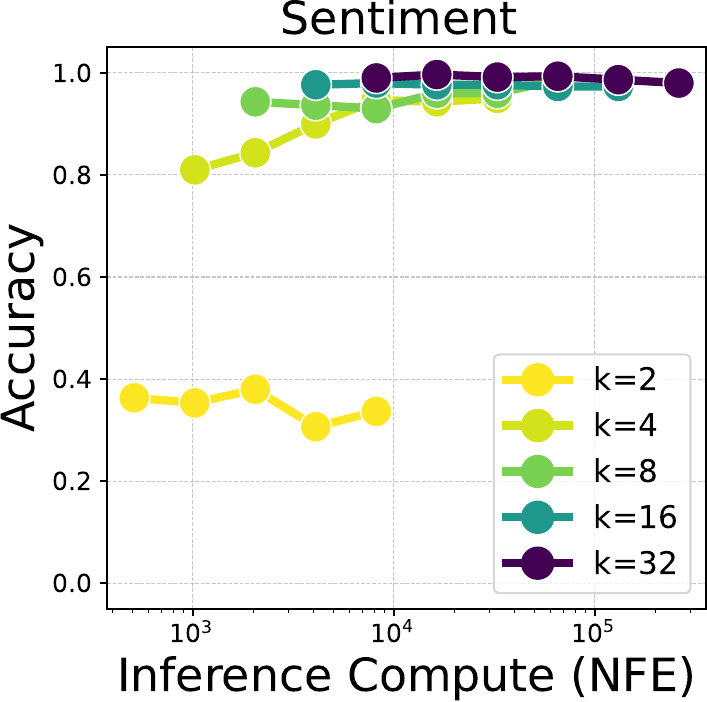}
        \caption{MDLM ($T$ via x-axis)}
        \label{fig:mdlm-sentiment}
    \end{subfigure}
    \caption{\label{fig:scale-particle}\footnotesize{Trade-offs between sample counts $k$ and denoising steps $T$ across compute budgets (NFEs). For (a) LLaDA, the x-axis shows NFEs controlled by varying $k$, with $T$ in the legend; for (b-d) MDLM, the x-axis shows NFEs controlled by varying $T$, with $k$ in the legend. Scaling $k$ (and decreasing $T$ accordingly) generally yields better performance under the same NFEs.}}}
    
\end{figure*}

\begin{table*}[h]
\centering
\caption{\footnotesize{Effective sample size (ESS) for \method{} across different numbers of iterations $m$ and particles per iteration $k$, under a fixed compute budget $m \times k = 64$.
}}
\label{tab:ess}
\setlength{\tabcolsep}{5.4pt}  %
\scriptsize
\begin{tabular}{lcccccccc}
\toprule
Setting & Iter 1 & Iter 2 & Iter 3 & Iter 4 & Iter 5 & Iter 6 & Iter 7 & Iter 8 \\
\midrule
$m{=}1$, $k{=}64$ & 60.2 $\pm$ 5.3 & -- & -- & -- & -- & -- & -- & -- \\
$m{=}2$, $k{=}32$ & 29.0 $\pm$ 4.1 & 30.6 $\pm$ 3.1 & -- & -- & -- & -- & -- & -- \\
$m{=}4$, $k{=}16$ & 13.3 $\pm$ 3.0 & 14.9 $\pm$ 2.1 & 15.2 $\pm$ 1.9 & 15.5 $\pm$ 1.2 & -- & -- & -- & -- \\
$m{=}8$, $k{=}8$  & 5.6 $\pm$ 1.9 & 6.8 $\pm$ 1.8 & 7.2 $\pm$ 1.5 & 7.5 $\pm$ 1.3 & 7.6 $\pm$ 0.9 & 7.7 $\pm$ 0.8 & 7.8 $\pm$ 0.5 & 7.8 $\pm$ 0.6 \\
\bottomrule
\end{tabular}
\end{table*}
\vspace{-0.5em}
\paragraph{Denoising Steps vs. Sample Count.}
We fix $T$ in prior experiments and now study the trade-off between $T$ and $k$. While $T \approx L$ is typically sufficient for DLMs~\citep{sahoo2024simple}, \method{} performs reward computation and resampling at every step, so larger $T$ can better concentrate samples. Should we then prioritize increasing $T$ or $k$? We fix $L=128$ and vary both: for LLaDA, we decrease $T$ (from 128 to 64, 32) while increasing $k$; for MDLM, we vary $T$ from 128 to 2048 and $k$ from 2 to 32, maintaining constant NFEs. As shown in Figure~\ref{fig:scale-particle}, increasing $k$ generally provides greater benefits, though when performance saturates, smaller $k$ can be preferable (Figure~\ref{fig:mdlm-cola}). This trend holds across Best-of-$n$ and vanilla SMC (Appendix~\ref{app:scale-tk}).

\section{Conclusion}
We introduced PG-DLM, a particle Gibbs sampling algorithm for DLMs that enables trajectory-level refinement at inference time. By constructing a Markov chain over complete denoising trajectories, PG-DLM introduces a new scaling axis, the number of refinement iterations, unavailable to single-pass methods. Future directions include characterizing the mixing rate of the particle Gibbs chain for tighter variance bounds and applying PG-DLM to other discrete generative tasks such as molecule design.

\section*{Limitations}
\method{} has three main limitations. First, in the low-particle, fixed-iteration regime,
\method{}~(fixed) can underperform SMC. Second, the primary gains come from iteration scaling
($m>1$), so tasks already near saturation at $m=1$, such as the controllable-generation tasks
in Tables~\ref{tab:control-gen-llada} and~\ref{tab:control-gen-mdlm}, offer limited headroom.
Third, as with other inference-time scaling methods, performance remains bounded by the capabilities of the base model.

\section*{Ethics Statement}
This paper presents work whose goal is to advance the field of Machine
Learning. There are many potential societal consequences of our work. We highlight that
\chg{controllable generation methods can be used to align models with human preferences. Additionally, we recognize that these methods can be used for automated red-teaming, which, if misused, could be used to generate harmful or unsafe content. However, we believe publishing these methods in a transparent and reproducible way enables the research community to better understand behaviors of generative models and develop stronger safeguards. We believe the benefits of this understanding will ultimately outweigh potential risks.}

\section*{Code Availability}
Code is available at \url{https://github.com/MhDang/pg-dlm}.

\section*{Acknowledgments}
This research is supported in part by ARO (W911NF-21-1-0125), ONR (N00014-23-1-2159), the CZ Biohub, and IBM.

\bibliography{main}
\bibliographystyle{colm2026_conference}

\appendix
\newpage

\section{Proof}
\label{app:proof}
\subsection{Optimal Denoising Distribution (Equation~\ref{eq:p_cond})}
\label{app:smc-optimal}

Following~\citet{uehara2024bridging,uehara2024fine}, we derive the reward-weighted conditional $p^*(\rvx_{t-1} \given \rvc, \rvx_t)$ from a per-step KL-regularized RL objective. Define the partial reward $r(\rvc, \rvx_t)$ as the expected future reward at timestep $t$:
\begin{equation}
    r(\rvc, \rvx_t) = \beta \log \E_{\rvx_0 \sim p_\theta(\rvx_0 \given \rvc, \rvx_t)} \left[ \exp\left( r(\rvc, \rvx_0)/\beta \right) \right].
    \label{eq:r_xt}
\end{equation}
The optimal conditional maximizes expected partial reward while staying close to the base denoiser:
\begin{equation}
    p^*(\rvx_{t-1} \given \rvc, \rvx_t) = \argmax_p \E_p \left[ r(\rvc, \rvx_{t-1}) \right] - \beta \KL \left[ p(\rvx_{t-1} \given \rvc, \rvx_t) \,\|\, p_\theta(\rvx_{t-1} \given \rvc, \rvx_t) \right].
    \label{eq:p_star_objective}
\end{equation}
The solution is tractable:
\begin{equation}
    p^*(\rvx_{t-1} \given \rvc, \rvx_t) \propto p_\theta(\rvx_{t-1} \given \rvc, \rvx_t) \exp\left( r(\rvc, \rvx_{t-1})/\beta \right).
    \label{eq:app-optimal-1}
\end{equation}
Normalizing yields:
\begin{align}
    p^*(\rvx_{t-1} \given \rvc, \rvx_t) &= \frac{p_\theta(\rvx_{t-1} \given \rvc, \rvx_t) \exp\left( r(\rvc, \rvx_{t-1})/\beta \right)}{\sum_{\rvx'_{t-1}} p_\theta(\rvx'_{t-1} \given \rvc, \rvx_t) \exp\left( r(\rvc, \rvx'_{t-1})/\beta \right)} \label{eq:p_cond_line2} \\
    &= p_\theta(\rvx_{t-1} \given \rvc, \rvx_t) \exp\left( \frac{r(\rvc, \rvx_{t-1}) - r(\rvc, \rvx_t)}{\beta} \right),
\end{align}
where the denominator from Equation~\ref{eq:p_cond_line2} equals $\exp\left( r(\rvc, \rvx_t)/\beta \right)$ by the soft Bellman equation (Theorem 1 of~\citet{uehara2024bridging}):
\[
r(\rvc, \rvx_t) = \beta \log \sum_{\rvx_{t-1}} p_\theta(\rvx_{t-1} \given \rvc, \rvx_t) \exp\left( r(\rvc, \rvx_{t-1})/\beta \right).
\]
This yields Equation~\ref{eq:p_cond}, decomposing the global RL objective (Equation~\ref{eq:rl}) across timesteps.

\subsection{Proof of the Variance Bound (Theorem~\ref{the:variance})}
\label{app:variance-proof}

Assume partial reward estimation is accurate as $\phi \to \infty$ (Remark~\ref{rem:partial_reward}). Abusing notation, we suppress the fixed conditioning prompt $\rvc$ (e.g., $p_\theta(\rvx_0) \equiv p_\theta(\rvx_0 \given \rvc)$). Let the proposal be the base model $p_\theta(\rvx_{0:T}) = p_\theta(\rvx_T) \prod_{t=1}^T p_\theta(\rvx_{t-1} \given \rvx_t)$, and define the reweighting function $\gamma(\rvx_0) = \exp(r(\rvx_0)/\beta)$.

The unnormalized target is then
\[
\tilde{p}(\rvx_{0:T}) = \gamma(\rvx_0) p_\theta(\rvx_{0:T}),
\]
with normalizing constant
\[
Z = \sum_{\rvx_{0:T}} \tilde{p}(\rvx_{0:T}) = \sum_{\rvx_{0:T}} \gamma(\rvx_0) p_\theta(\rvx_{0:T}) = \E_{p_\theta(\rvx_0)}[\gamma(\rvx_0)].
\]
The normalized target is $\pi(\rvx_{0:T}) = \tilde{p}(\rvx_{0:T}) / Z = \gamma(\rvx_0) p_\theta(\rvx_{0:T}) / Z$, which is essentially $p^*(\rvx_{0:T})$.

From~\citet{andrieu2010particle}, particle Gibbs variance is bounded by that of the underlying SMC. From~\citet{robert1999monte,chatterjee2018sample}, for the SMC estimator $\widehat{Z}$ with $N$ particles over trajectories $\rvx_{0:T}$ with proposal $p_\theta(\rvx_{0:T})$ and target $\pi(\rvx_{0:T})$,
\[
\mathrm{Var}(\widehat{Z}) \leq \frac{Z^2}{N} \left( \exp\left(\KL(\pi \| p_\theta)\right) - 1 \right),
\]
where $\pi$ and $p_\theta$ are defined over $\rvx_{0:T}$ and $N$ the number of particles. Now,
\[
\KL(\pi \| p_\theta) = \E_\pi\left[ \log \frac{\pi}{p_\theta}\right] =\E_\pi \left[\log \frac{\gamma(\rvx_0)}{Z} \right].
\]
By Jensen's inequality,
\[
\KL(\pi \| p_\theta) \leq \log \frac{\E_\pi\left[\gamma(\rvx_0)\right]}{Z} =\log \frac{\E_{p_\theta} [\gamma(\rvx_0)^2]}{Z^2}= \log \frac{\E_{p_\theta(\rvx_0)} [\gamma(\rvx_0)^2]}{Z^2}.
\]
Thus,
\begin{multline*}
\mathrm{Var}(\widehat{Z}) \leq \frac{Z^2}{N} \left( \frac{\E_{p_\theta(\rvx_{0})} [\gamma( \rvx_{0})^2]}{Z^2} - 1 \right) \\
= \frac{\E_{p_\theta(\rvx_{0})} [\gamma( \rvx_{0})^2]-\left(\E_{p_\theta(\rvx_{0})} [\gamma( \rvx_{0})]\right)^2}{N} = \frac{\mathrm{Var}_{p_\theta(\rvx_0)}(\gamma(\rvx_0))}{N}.
\end{multline*}
For \method{} with $k$ samples within a single iteration, $N=k$, yielding the stated bound. By~\citet{andrieu2010particle}, the variance of the particle Gibbs estimator is bounded by that of the underlying SMC estimator, completing the proof.

\section{Sequential Monte Carlo Background}
\label{app:is-smc}

\textbf{Importance Sampling (IS).}
To estimate expectations under a target $f(\rvx)$ (hard to sample from) using a proposal $g(\rvx)$ (easy to sample):
\[
\E_f[h(\rvx)] = \E_g \left[ h(\rvx) \frac{f(\rvx)}{g(\rvx)} \right] \approx \sum_{i=1}^N w_i h(\rvx^{(i)}), \quad \text{where} \quad w_i = \frac{f(\rvx^{(i)})}{g(\rvx^{(i)})}, \{\rvx^{(i)}\}_{i=1}^N \sim g.
\]
Resample with replacement via normalized $\{w_i\}$ for approximate samples from $f$.

\textbf{Sequential Importance Sampling (SIS).}
For sequential targets $f(\rvx) = \prod_t f(x_t \mid \rvx_{t-1})$ and proposals $g(\rvx) = \prod_t g(x_t \mid \rvx_{t-1})$, where the full variable is \(\rvx = (x_1, \dots, x_d)\) and partial prefix \(\rvx_t = (x_1, \dots, x_t)\) (with \(\rvx_0\) empty), weights factorize recursively:
\[
w_t(\rvx_t) = w_{t-1}(\rvx_{t-1}) \cdot \frac{f(x_t \mid \rvx_{t-1})}{g(x_t \mid \rvx_{t-1})}, \quad w_0 = 1.
\]
Propagate $x_t^{(i)} \sim g(\cdot \mid \rvx_{t-1}^{(i)})$, update $w_t^{(i)}$.

\textbf{Sequential Monte Carlo (SMC).}
SMC adds resampling to SIS to counter degeneracy. For $N$ particles $\{\rvx_t^{(i)}, w_t^{(i)}\}_{i=1}^N$:
\begin{enumerate}
\item Initialize $w_0^{(i)}=1$.
\item For $t=1,\dots,d$:
\begin{enumerate}
\item Propagate: $x_t^{(i)} \sim g(\cdot \mid \rvx_{t-1}^{(i)})$.
\item Weight: $\tilde{w}_t^{(i)} = w_{t-1}^{(i)} \cdot \frac{f(x_t^{(i)} \mid \rvx_{t-1}^{(i)})}{g(x_t^{(i)} \mid \rvx_{t-1}^{(i)})}$.
\item Resample $N$ indices $\propto$ normalized $\{\tilde{w}_t^{(i)}\}$; reset to equal weights.
\end{enumerate}
\end{enumerate}

\newpage
\section{Reward Functions and Baselines for Experiments}

We evaluate three reward functions for controllable generation:

\begin{enumerate}    
    \item \textbf{Linguistic Acceptability}: Favors grammatically correct sentences using a RoBERTa classifier~\citep{morris2020textattack} trained on CoLA~\citep{warstadt2019neural}. We measure CoLA classification accuracy. Model: \url{https://huggingface.co/textattack/roberta-base-CoLA}.
    
    \item \textbf{Controlled Toxicity}: Guides toward (or away from) toxic outputs using a RoBERTa toxicity classifier~\citep{logacheva2022paradetox} for red-teaming. We measure toxicity classification accuracy. Model: \url{https://huggingface.co/SkolkovoInstitute/roberta_toxicity_classifier}.
    
    \item \textbf{Controlled Sentiment}: Steers toward target sentiments (e.g., positive) using a RoBERTa classifier~\citep{barbieri2020tweeteval} on TweetEval. We measure sentiment classification accuracy. Model: \url{https://huggingface.co/cardiffnlp/twitter-roberta-base-sentiment}.

\end{enumerate}

Baseline implementations for FK-Steering and Best-of-$n$ are adapted from \url{https://github.com/zacharyhorvitz/Fk-Diffusion-Steering/tree/main/discrete_diffusion}; we re-ran experiments for consistency.

\section{Partial Reward Estimation}
To estimate partial rewards $r(\rvc, \rvx_t)$ for prompt $\rvc$ and noisy state $\rvx_t$, in order to compute importance weights (line 10 in Algorithm~\ref{alg:pg}), we approximate the expectation $\mathbb{E}_{p_\theta(\rvx_0 \mid \rvc, \rvx_t)} \left[ \exp\left( r(\rvc, \rvx_0)/\beta \right) \right]$ as in Equation~\ref{eq:p_cond} using $\phi$ samples $\rvx_0 \sim p_\theta(\rvx_0 \mid \rvc, \rvx_t)$ \chg{by unrolling $\tau$ diffusion steps per sample. In practice, we set $\tau=1$ for efficiency following prior work. However, studying the scaling behavior of $\tau$ is an interesting and promising complementary future direction.}

\begin{figure}[h]
    \centering
        \includegraphics[width=0.24\textwidth]{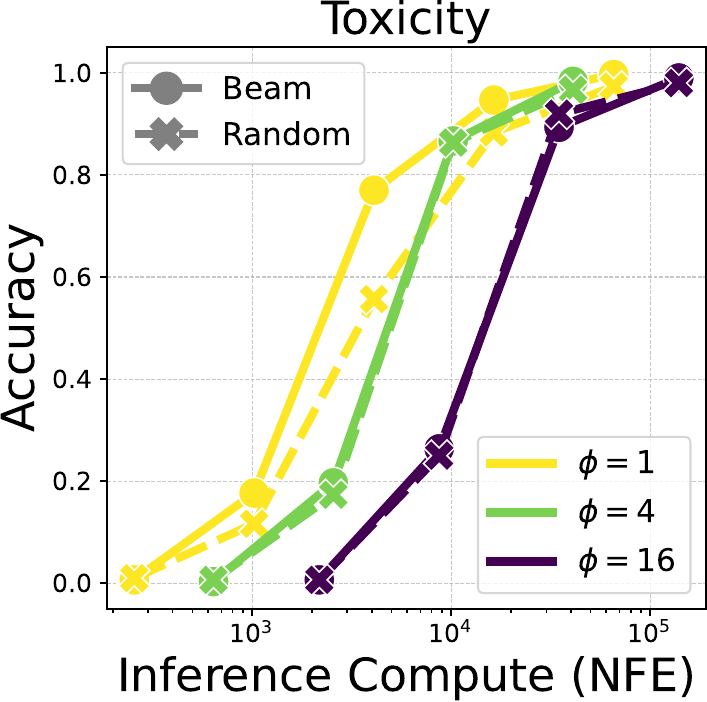}
        \includegraphics[width=0.24\textwidth]{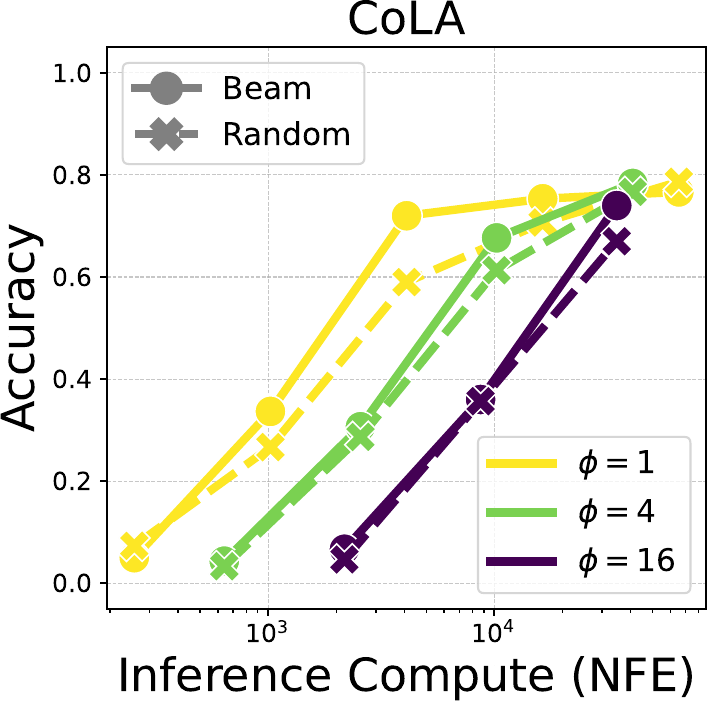}
        \includegraphics[width=0.24\textwidth]{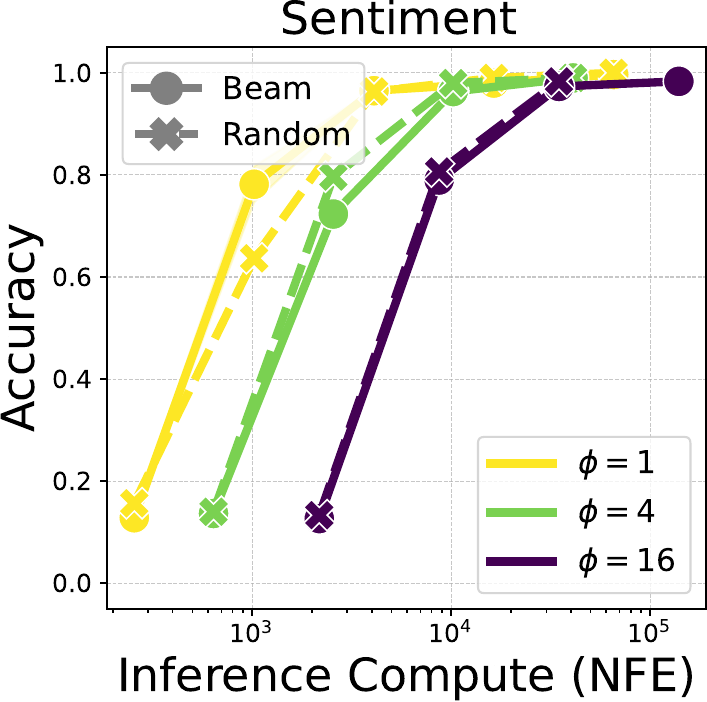}
        \includegraphics[width=0.24\textwidth]{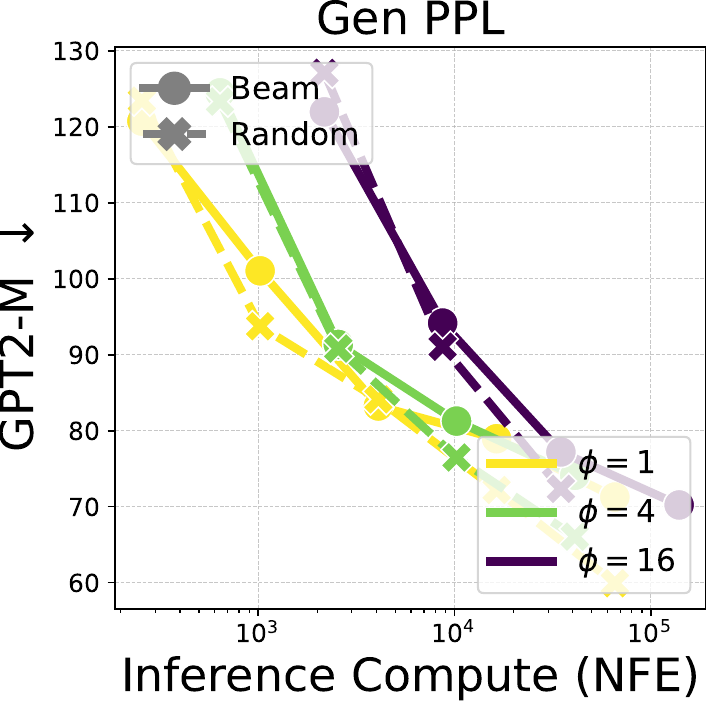}
        \caption{\label{fig:scale-reward-complete} Comparison of Beam and Random sampling for partial reward estimation with varying number of $\x0$ samples ($\phi$) across NFEs (as controlled by the number of samples $k$). Beam sampling with $\phi=1$ performs the best. }
\end{figure}

A common approach is to draw random samples from $p_\theta(\rvx_0 \mid \rvc, \rvx_t)$, yielding unbiased but high-variance estimates~\citep{singhal2025fksteering, song2021scorebased, wu2023practical, li2024derivative}.
We instead propose \emph{beam sampling} to approximate $p_\theta(\rvx_0 \mid \rvc, \rvx_t)$, with $\phi$ as the beam width, yielding biased but low-variance estimates. For $\phi=1$, this reduces to greedy decoding. As shown in Figure~\ref{fig:scale-reward-complete}, scaling $\phi$ improves accuracy but raises compute, leading to suboptimal trade-offs. Beam sampling outperforms random methods in most cases, with $\phi=1$ offering the best trade-off.

\newpage
\section{Additional Experimental Results}
\subsection{m vs k Scaling}

Table~\ref{tab:scaling-results} shows detailed controlled text performance across reward functions (CoLA, Toxicity, Sentiment) under varying compute budgets $n=mk$ on MDLM with different particle Gibbs iterations $m$ and sample counts $k$, fixing generated length $L=128$ and $T=128$. Each row fixes $n$ while varying $m$ and $k$; best per row bolded. At higher $n$, increasing $k$ can yield diminishing returns, in which case scaling $m$ becomes more effective, though the optimal allocation depends on the reward function.

\begin{table}[ht]
\centering
\caption{\footnotesize{Controlled text performance across reward functions under varying compute budgets, with different \(m\) and \(k\). Best per row bolded. Note: generation length and denoising steps ($L{=}128$, $T{=}128$) differ from Table~\ref{tab:control-gen-mdlm} ($L{=}50$, $T{=}1024$), so accuracies are not directly comparable across the two tables.}}
\label{tab:scaling-results}
\footnotesize
\begin{tabular}{clclclclc}
\toprule
\multirow{2}{*}{Metric} & \multicolumn{2}{c}{$m=1$} & \multicolumn{2}{c}{$m=2$} & \multicolumn{2}{c}{$m=4$} & \multicolumn{2}{c}{$m=8$} \\
\cmidrule(lr){2-3} \cmidrule(lr){4-5} \cmidrule(lr){6-7} \cmidrule(lr){8-9}
& $k$ & Accuracy & $k$ & Accuracy & $k$ & Accuracy & $k$ & Accuracy \\
\midrule
\multirow{5}{*}{CoLA $\uparrow$} 
& 16 & 87.3 & 8 & 87.0 & 4 & \textbf{89.7} & 2 & 79.0 \\
& 32 & 89.7 & 16 & 84.0 & 8 & 88.7 & 4 & \textbf{90.0} \\
& 64 & 85.7 & 32 & 79.7 & 16 & 86.3 & 8 & \textbf{88.7} \\
& 128 & \textbf{86.3} & 64 & 79.0 & 32 & 83.3 & 16 & 80.3 \\
& 256 & 78.7 & 128 & \textbf{80.0} & 64 & 73.0 & 32 & 77.0 \\
\midrule
\multirow{5}{*}{Toxicity $\uparrow$} 
& 16 & \textbf{81.3} & 8 & 73.7 & 4 & 59.0 & 2 & 15.7 \\
& 32 & 90.3 & 16 & \textbf{93.7} & 8 & 91.7 & 4 & 78.3 \\
& 64 & 96.3 & 32 & 97.0 & 16 & \textbf{97.7} & 8 & \textbf{97.7} \\
& 128 & 98.7 & 64 & \textbf{99.7} & 32 & 98.3 & 16 & 98.0 \\
& 256 & 98.7 & 128 & 99.0 & 64 & \textbf{99.7} & 32 & 99.3 \\
\midrule
\multirow{5}{*}{Sentiment $\uparrow$} 
& 16 & 97.7 & 8 & \textbf{99.0} & 4 & 98.0 & 2 & 82.7 \\
& 32 & 99.0 & 16 & 99.7 & 8 & \textbf{100.0} & 4 & 99.0 \\
& 64 & 99.7 & 32 & \textbf{100.0} & 16 & 99.7 & 8 & 98.7 \\
& 128 & \textbf{100.0} & 64 & 99.7 & 32 & 99.7 & 16 & 99.7 \\
& 256 & 99.3 & 128 & 99.7 & 64 & \textbf{100.0} & 32 & 99.7 \\
\bottomrule
\end{tabular}
\end{table}

\newpage

\subsection{T vs k Scaling: Additional Results for Figure~\ref{fig:scale-particle}}
\label{app:scale-tk}
Figure~\ref{fig:scale-particle} illustrates trade-offs between sample counts and denoising steps for \method{}. Here we show the same trend holds for baselines: FK-Steering~\citep{singhal2025fksteering} and Best-of-$n$, where scaling samples generally outperforms scaling denoising steps under the same compute budget. We use MDLM as the base model.

\begin{enumerate}
    \item For SMC with number of $\rvx_0$ samples $\phi=1$: 
\begin{figure}[h]
    \centering
    \begin{minipage}[b]{0.24\columnwidth}
        \centering
        \includegraphics[width=1.0\linewidth]{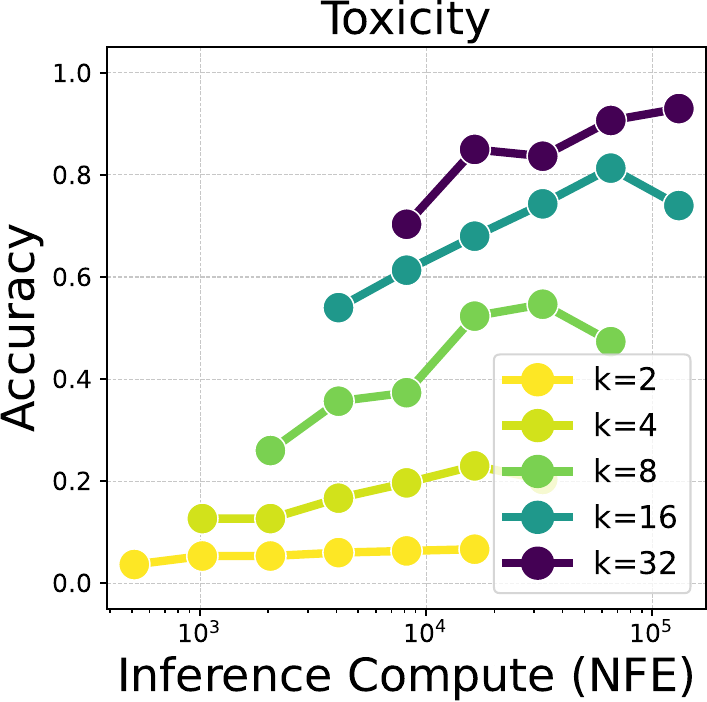}
    \end{minipage}
    \begin{minipage}[b]{0.24\columnwidth}
        \centering
        \includegraphics[width=1.0\linewidth]{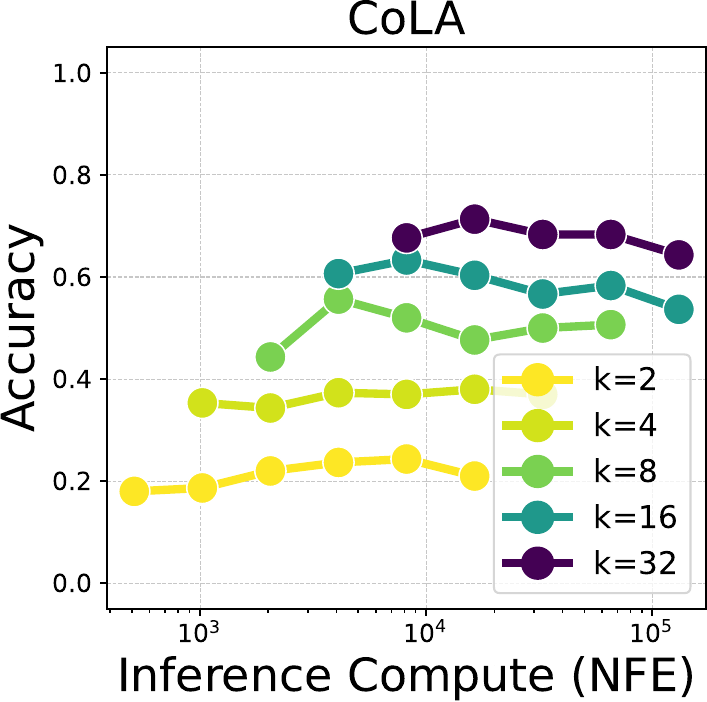}
    \end{minipage}
    \begin{minipage}[b]{0.24\columnwidth}
        \centering
        \includegraphics[width=1.0\linewidth]{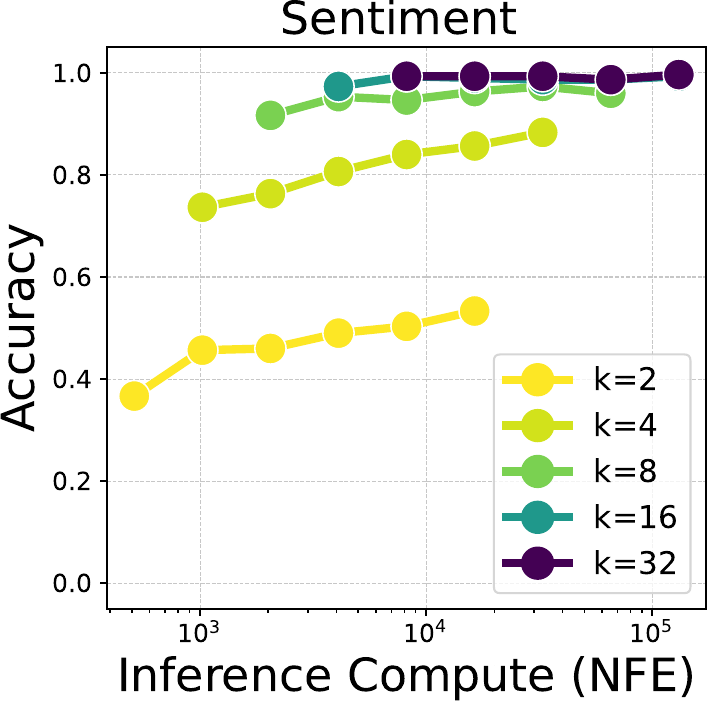}
    \end{minipage}
    \begin{minipage}[b]{0.24\columnwidth}
        \centering
        \includegraphics[width=1.0\linewidth]{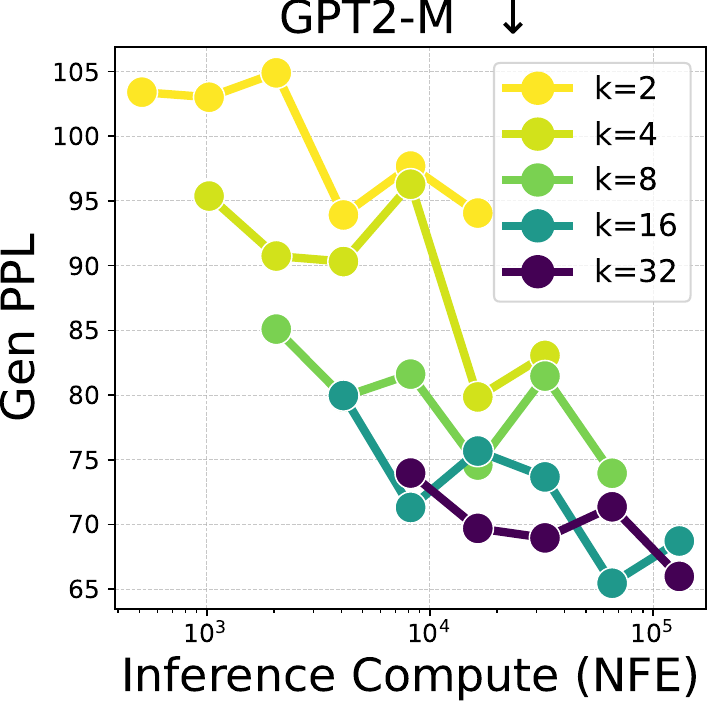}
    \end{minipage}
    \caption{\label{fig:scale-particle-smc1} \footnotesize{Trade-offs between sample counts $k$ and denoising steps $T$ across compute budgets (NFEs) for \textbf{SMC ($\phi=1$)}. The x-axis shows NFEs controlled by varying $T$, with $k$ in the legend. Scaling $k$ (and decreasing $T$ accordingly) generally
yields better performance under the same NFEs.}
    }
\end{figure}

\item For SMC with number of $\rvx_0$ samples $\phi=4$:
\begin{figure}[h]
    \centering
    \begin{minipage}[b]{0.24\columnwidth}
        \centering
        \includegraphics[width=1.0\linewidth]{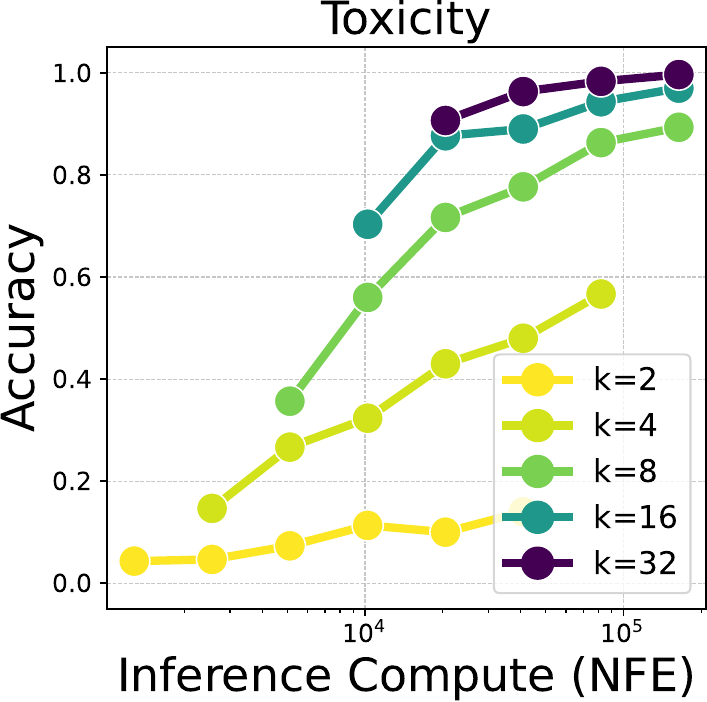}
    \end{minipage}
    \begin{minipage}[b]{0.24\columnwidth}
        \centering
        \includegraphics[width=1.0\linewidth]{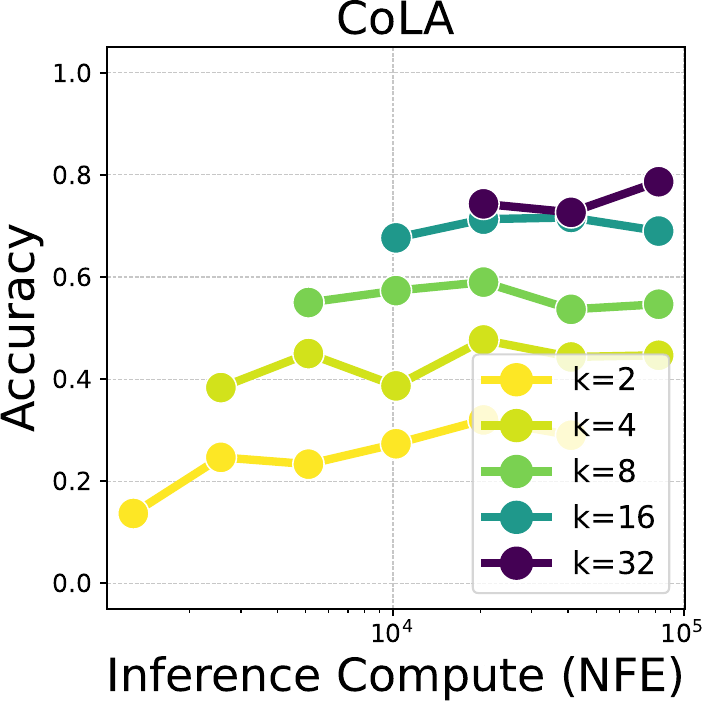}
    \end{minipage}
    \begin{minipage}[b]{0.24\columnwidth}
        \centering
        \includegraphics[width=1.0\linewidth]{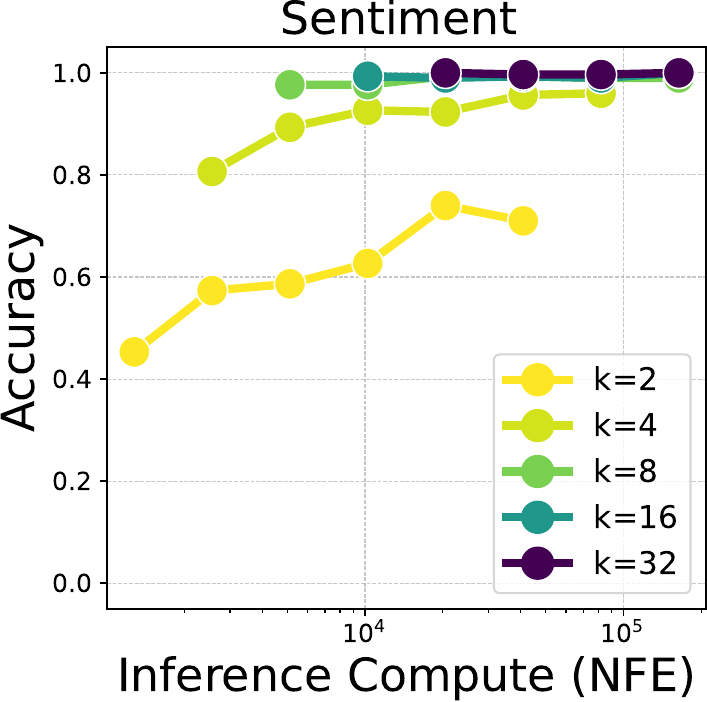}
    \end{minipage}
    \begin{minipage}[b]{0.24\columnwidth}
        \centering
        \includegraphics[width=1.0\linewidth]{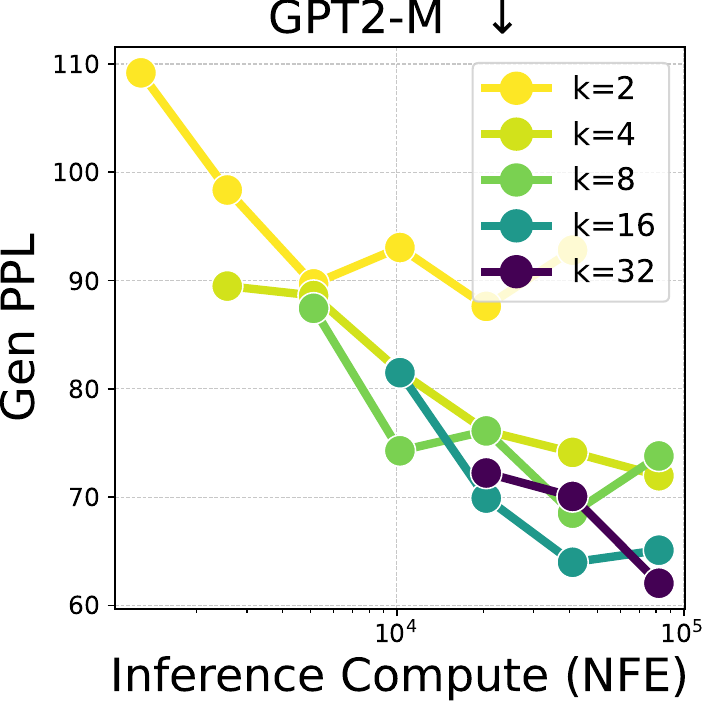}
    \end{minipage}
    \caption{\label{fig:scale-particle-smc4}  \footnotesize{Trade-offs between sample counts $k$ and denoising steps $T$ across compute budgets (NFEs) for \textbf{SMC ($\phi=4$)}. The x-axis shows NFEs controlled by varying $T$, with $k$ in the legend. Scaling $k$ (and decreasing $T$ accordingly) generally
yields better performance under the same NFEs.}}
\end{figure}
\item For Best-of-$n$:
\begin{figure}[h]
    \centering
    \begin{minipage}[b]{0.24\columnwidth}
        \centering
        \includegraphics[width=1.0\linewidth]{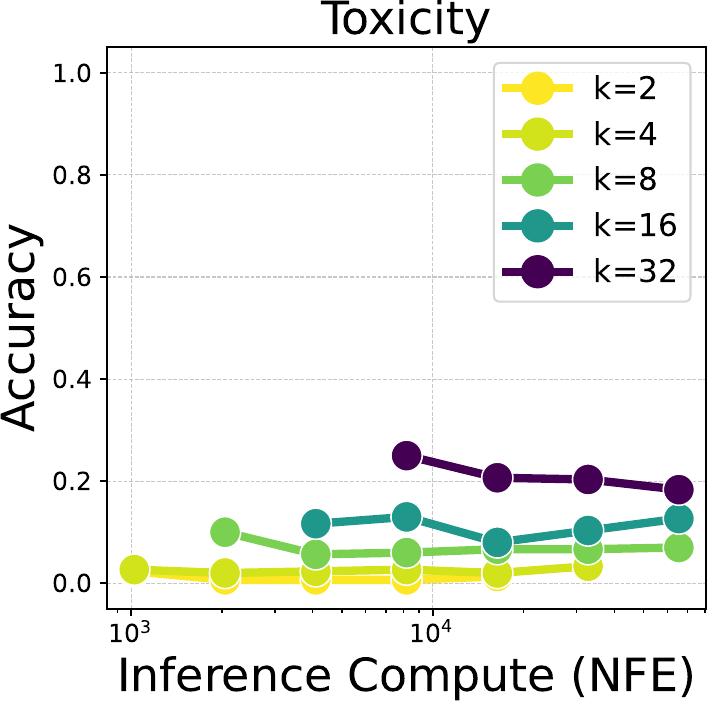}
    \end{minipage}
    \begin{minipage}[b]{0.24\columnwidth}
        \centering
        \includegraphics[width=1.0\linewidth]{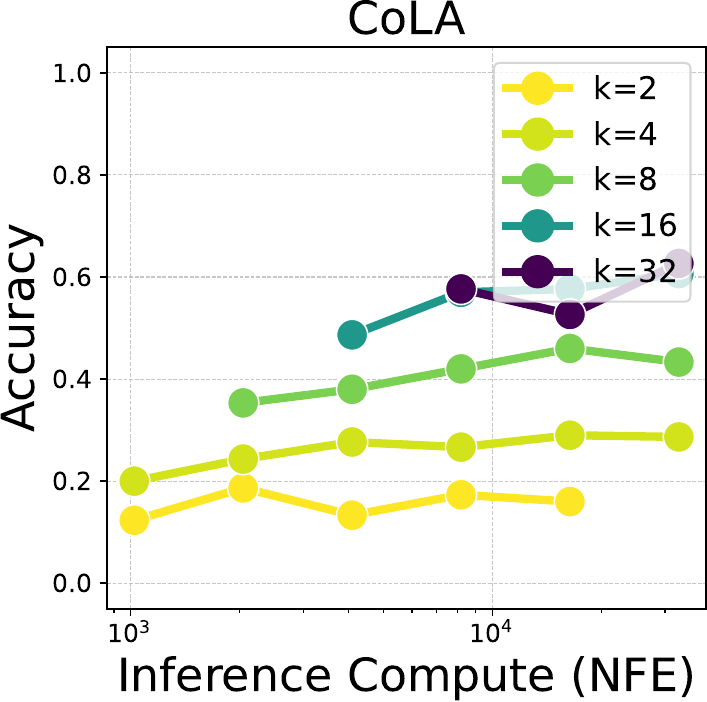}
    \end{minipage}
    \begin{minipage}[b]{0.24\columnwidth}
        \centering
        \includegraphics[width=1.0\linewidth]{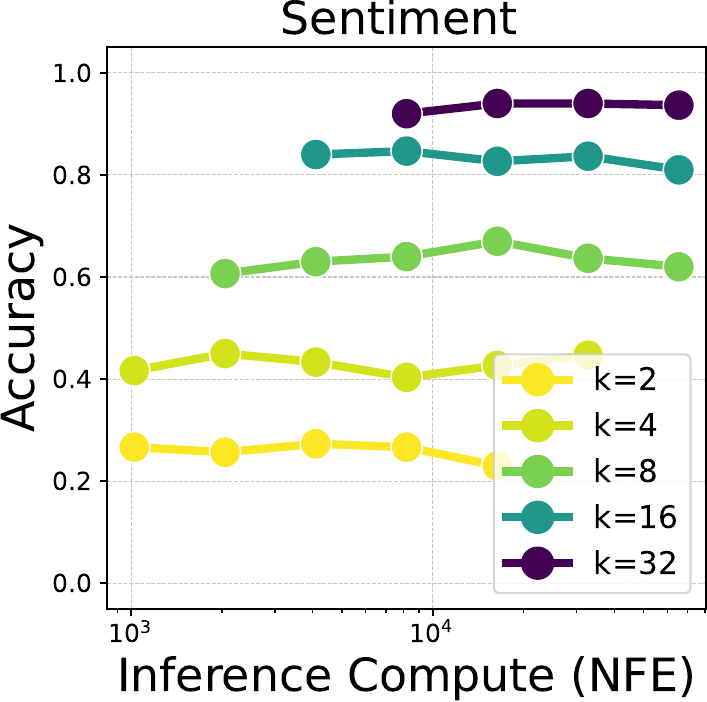}
    \end{minipage}
    \begin{minipage}[b]{0.24\columnwidth}
        \centering
        \includegraphics[width=1.0\linewidth]{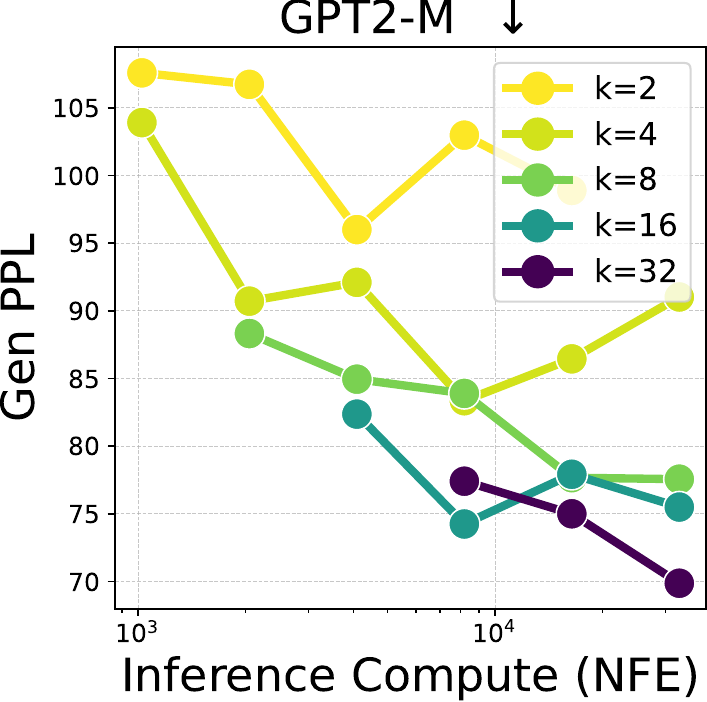}
    \end{minipage}
    \caption{\label{fig:scale-particle-bon} \footnotesize{Trade-offs between sample counts $k$ and denoising steps $T$ across compute budgets (NFEs) for \textbf{Best-of-$n$}. The x-axis shows NFEs controlled by varying $T$, with $k$ in the legend. Scaling $k$ (and decreasing $T$ accordingly) generally
yields better performance under the same NFEs. }}
\end{figure}
\end{enumerate}

\newpage
\section{\chg{Qualitative Examples}}
\begin{table}[htbp]
\centering
\begin{tabular}{@{}p{2cm}p{11cm}@{}}
\toprule
\textbf{Method} & \textbf{Generated Output} \\
\midrule
Best-of-$n$ & \begin{itemize}[leftmargin=1em]
    \item Once upon a time, this was one of my favorite taglines in Indie Match Match :The impossible we overcome Those that we escape The Impossible were our face. The Impossible were our face
    \item The chicken is still really amazing after consuming the amount is parox Imagine had orange soup. The soup has very low sugar release. The whole concept of this is that it helps as an antioxidant. It's an antioxidan
    \item The lake went up through the fields, the hills cracked, and fell to the sea. Heaven came clean, the wind sang like the mountains: BRAND BLOOD Now black, skin on cold, Ice white
\end{itemize}\\
FK & 
\begin{itemize}[leftmargin=1em]
  \item Once upon a time,was one of the coolest and most beautiful colors of all time. Nowadays, this color is among my favorite colors of all time. Let me show you guys with some pictures of what my favorite colors look like
  \item The chicken was extremely tender and flavorful. There was a nice crunchiness to chicken wings on top. I do prefer to eat chicken wings when they are a little smaller and less crunchy. I also enjoyed keeping the wings in the refrigerator
  \item The lake temperature is colder in the spring, which allows you to use the water easier. At a depth above the current lake level, you can find the most beautiful thermal lakes in North America. The lakes are brilliant
\end{itemize}\\
PG-DLM & 
\begin{itemize}[leftmargin=1em]
\item Once upon a time, the openmindedness and diversity of the universe was one of the pillars of our success, and continues to be. Today, we welcome the diversity and nature of the universe, and embrace it as a
\item The chicken burger really live up to the deli's spot for the dish. The fried chicken wings really make it an addition of the menu due to their cute goo and I LOVE THEM! The burger isn't the best 
\item  The lake itself is totally potable and there are plenty of holes in the middle of the lake. It is perfect for any kind of tradition of mountaineering adventure.The lake is also used as a point of contact and
\end{itemize}
\\
\bottomrule
\end{tabular}
\caption{Qualitative comparison of generated sequences under a positive sentiment reward}
\label{tab:qualitative_language}
\end{table}
\paragraph{Use of AI Tools.}
We used large language models (e.g., ChatGPT) to assist with minor language editing and improving clarity of the manuscript. All technical content, experiments, and conclusions are solely the work of the authors.

\end{document}